\documentclass[a4paper,fleqn]{cas-dc}

\usepackage[authoryear]{natbib}
\usepackage{amsthm,amsmath,amssymb}
\usepackage{algorithm}
\usepackage{algorithmic}
\usepackage{graphicx}
\usepackage{float}
\usepackage{subfigure}
\usepackage{caption}
\usepackage{threeparttable}
\usepackage{color}
\usepackage{booktabs}
\usepackage{multirow}
\usepackage{array}
\usepackage{enumitem}

\makeatletter
\renewcommand{\fnum@figure}{Fig. \thefigure.\@gobble }
\makeatother

\def\tsc#1{\csdef{#1}{\textsc{\lowercase{#1}}\xspace}}
\tsc{WGM}
\tsc{QE}
\tsc{EP}
\tsc{PMS}
\tsc{BEC}
\tsc{DE}

\begin{document}
\begin{sloppypar}
\let\printorcid\relax
\let\WriteBookmarks\relax
\def\floatpagepagefraction{1}
\def\textpagefraction{.001}

\shorttitle{}

\shortauthors{Xinzheng Wu et~al.}

\title [mode = title]{Make Full Use of Testing Information: An Integrated Accelerated Testing and Evaluation Method for Autonomous Driving Systems}                      



%
\author[1]{\textcolor{black}{Xinzheng Wu}}[]
\credit{Formal analysis, Methodology, Software, Visualization, Writing - original draft}
\affiliation[1]{organization={School of Automitive Studies, Tongji University},
    addressline={No. 4800 Cao'an Road.}, 
    city={Shanghai},
    postcode={201804}, 
    country={China}}

\author[1]{\textcolor{black}{Junyi Chen}}[]
\credit{Formal analysis, Project administration, Validation, Writing - review \& editing}
\cormark[1]
\ead{chenjunyi@tongji.edu.cn}

\author[1]{\textcolor{black}{Jianfeng Wu}}[]
\credit{Conceptualization, Formal analysis, Writing - review \& editing}

\author[1]{\textcolor{black}{Longgao Zhang}}[]
\credit{Data curation, Software, Validation}

\author[1]{\textcolor{black}{Tian Xia}}[]
\credit{Investigation, Data curation, Validation}

\author[1]{\textcolor{black}{Yong Shen}}[]
\credit{Investigation, Resources, Supervision}

\cortext[cor1]{Corresponding author}



\begin{abstract}
Testing and evaluation is an important step before the large-scale application of the autonomous driving systems (ADSs). Based on the three level of scenario abstraction theory, a testing can be performed within a logical scenario, followed by an evaluation stage which is inputted with the testing results of each concrete scenario generated from the logical parameter space. During the above process, abundant testing information is produced which is beneficial for comprehensive and accurate evaluations. To make full use of testing information, this paper proposes an \textbf{I}ntegrated accelerated \textbf{T}esting and \textbf{E}valuation \textbf{M}ethod (ITEM). Based on a Monte Carlo Tree Search (MCTS) paradigm and a dual surrogates testing framework proposed in our previous work, this paper applies the intermediate information (i.e., the tree structure, including the affiliation of each historical sampled point with the subspaces and the parent-child relationship between subspaces) generated during the testing stage into the evaluation stage to achieve accurate hazardous domain identification. Moreover, to better serve this purpose, the UCB calculation method is improved to allow the search algorithm to focus more on the hazardous domain boundaries. Further, a stopping condition is constructed based on the convergence of the search algorithm. Ablation and comparative experiments are then conducted to verify the effectiveness of the improvements and the superiority of the proposed method. The experimental results show that ITEM could well identify the hazardous domains in both low- and high-dimensional cases, regardless of the shape of the hazardous domains, indicating its generality and potential for the safety evaluation of ADSs.
\end{abstract}



\maketitle

\section{Introduction}

The safety of autonomous driving systems (ADS) is a pivotal issue that demands comprehensive verification prior to the large-scale deployment \citep{sohrabi2021quantifying}. As a crucial aspect of safety, the safety of the intended functionality (SOTIF), as defined in the automotive safety standard ISO 21448 \citep{iso21448}, focuses on and endeavors to eliminate hazards or risks that caused by insufficiencies of specification or performance limitations of ADSs. According to the standard, in the verification phase, the SOTIF problem can be addressed by simulation testing to evaluate ADSs under known hazardous scenarios \citep{wang2024runtime}. Currently, based on the three level of scenario abstraction theory \citep{menzel2018scenarios}, scenario-based simulation testing and evaluation has become the prevailing verification method to address SOTIF problem due to its low cost, high efficiency and repeatability, which has garnered widespread attention from both academia and industry \citep{sun2022scenariobased}.

\begin{figure}[t] 
      \centering
      \includegraphics[width=8cm]{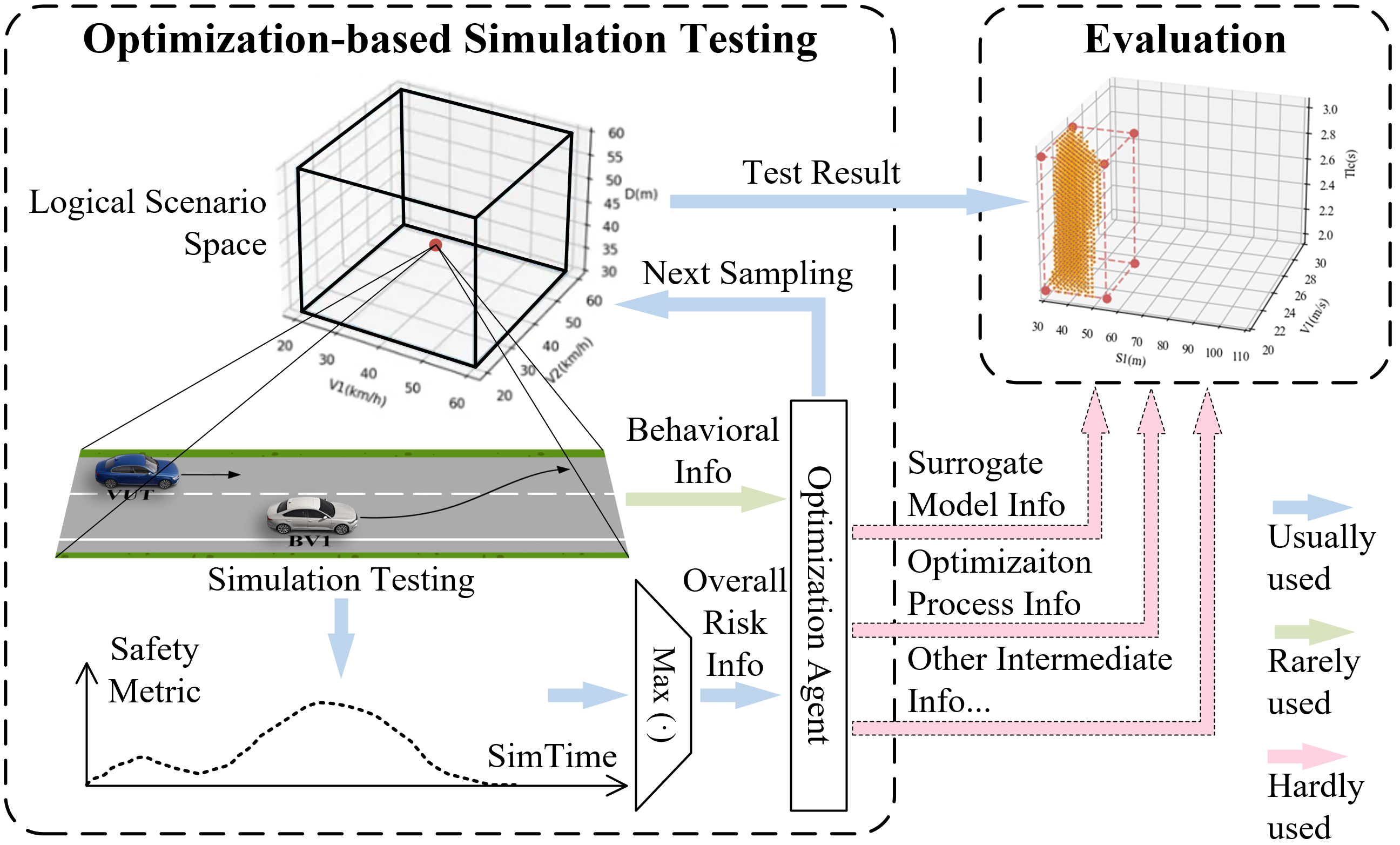}
      \caption{Information at different layers during the testing and evaluation process.}
      \label{info_layer}
\end{figure}

In practice, testing and evaluation are mainly conducted on the logical scenario level \citep{peixingzhang2022performance}. However, due to the high complexity and uncertainty of the external environment of high-level ADS, the logical scenario space constructed is usually high-dimensional, leading to the "dimensionality explosion" problem \citep{feng2023dense}. To address this issue, optimization algorithms are introduced by numerous researchers to effectively search hazardous scenarios in the whole logical scenario space \citep{zhang2023findinga}. Generally, as shown by the blue arrow in Fig. \ref{info_layer}, during the testing process, the risk of a test scenario for the vehicle under test (VUT) is calculated simultaneously using certain safety metrics \citep{wang2021review, lu2021performance}. After a test scenario is completed, an overall risk result, representing the safety performance of the VUT, can be obtained, which is then fed into the optimization agent as a cost function value. With a batch of test results obtained, the optimization agents will guide the sampling of the next batch of test scenarios towards the predicted possible hazardous areas. Finally, when the stopping condition of the optimization algorithm is satisfied, all the test results will be sent to the evaluation module for the safety verification of VUT.

Following the above technical route, existing methods have realized the accelerated generation of hazardous scenarios \citep{gong2023adaptive, huang2024bayesian}. However, in the above process, only scenario risk information is used, which hinders further improvement of testing efficiency and the comprehensiveness of evaluation. In fact, all other diverse information generated during the testing process can provide insight for the testing itself as well as the evaluation. As a remedy, our previous work \citep{wu2024accelerated} takes the behavioral information (as shown by the green arrow in Fig. \ref{info_layer}) into consideration. By using a convolutional neural network (CNN) to extract the information from the trajectories of vehicles, we have constructed a dual surrogates-based accelerated testing method and improved both the efficiency and coverage of the generation of hazardous scenarios. Nevertheless, testing and evaluation remain separate in the above work. Still, only the scenario risk information is transferred to the evaluation process, ignoring the intermediate information generated during the optimization process, as shown by the pink arrow in Fig. \ref{info_layer}. Towards addressing this issue, this paper establishes an Integrated accelerated Testing and Evaluation Method (ITEM) so as to make full use of the testing information. Compared with our previously published paper \citep{wu2024accelerated}, the new contributions of this paper are listed as follows:

\begin{enumerate}[label=(\arabic*)]
  \item An integrated testing and evaluation framework is proposed with the aim of leveraging testing information to obtain a more accurate and comprehensive evaluation result.
  \item In the accelerated testing stage, the UCB calculation method is improved to enable sampling to converge at the hazardous domain boundaries, thereby facilitating accurate evaluation. Additionally, a stopping condition is constructed based on the the idea of algorithmic convergence.
  \item In the accelerated evaluation stage, a hazardous domain identification method is proposed using both the testing results (i.e., sampling records) and the intermediate testing information (i.e., tree structure).
  \item Two metrics, namely API and ADI, are proposed quantify the accuracy of hazardous domain identification. These two metrics are designed with considerations from the perspectives of proportional accuracy and distributional accuracy respectively.
  \item Ablation and comparative experiments are conducted to verify the effectiveness of the improvements and the superiority of the proposed method compared with other baseline methods.
\end{enumerate}

The remainder of the paper is organized as follows: Some related works and research gaps are described in Sec. \ref{related_works}. In Sec. \ref{method}, the framework and the improvements of ITEM is detailed. Ablation and comparative experiments are performed in Sec. \ref{experiments}, followed by the analysis and discussion of the experimental results. The conclusion is summarized in Sec. \ref{conclusion}.

\section{Related Works} \label{related_works}

\subsection{Safety Testing Methods for Autonomous Driving Systems}

In the current literature, safety testing methods for ADS can be classified as Naturalistic Driving Data-based (NDD-based), Design of Experiments-based (DOE-based), and adaptive DOE-based (ADOE-based) methods.

By statistically analyzing the collected naturalistic driving data, NDD-based methods directly extract testing scenarios to satisfy specific testing requirement. For example, by translating the data from different sources into a standardized format and calculating the possibility of recombination between scenario slices, the project PEGASUS proposed a scenario extraction and classification method \citep{putz2017database}. Similarly, Yin et al. extracted hazardous lane-changing testing scenarios from China-FOT naturalistic data by establishing a scenario risk classification model and an excellent human driver model \citep{yin2023testa}. Furthermore, Feng et al. first extracted all cut-in events in a NDD database and then applied the seed-fill method to search hazardous scenarios \citep{feng2020safety}. Besides, Importance Sampling (IS) was also widely used to accelerate rare-scenario probability estimation, thus increasing the number of rare scenarios in NDD sampling \citep{zhao2017accelerated, feng2021testing}. Since the generated testing scenarios are derived from real driving data, NDD-based methods are of high realism. However, these methods rely on a large amount of high-quality data and the generated scenarios are limited to the known data, resulting in low generalizability. 

Instead of directly extracting testing scenarios from the collected data, DOE-based methods design testing scenarios by using different combinations of scenario parameters, where all testing scenarios are generated before the start of the test. Combinatorial testing is a typical representative of the DOE-based approaches. Bagschik et al. constructed a five-layer scenario model and generated functional scenarios based on parameter combinations \citep{bagschik2018ontology}. Li et al. described the environment of VUT through an ontology model, which was then used as input for combinatorial testing \citep{li2020ontologybased}. Apart from that, methods such as random testing, near-random testing (e.g. Latin Hypercube Sampling \citep{batsch2019performancea}) and grid testing can also be considered as DOE-based methods. DOE-based methods ignore the information of the VUT that is gradually acquired during the testing process, and the subsequent testing scenarios cannot be designed with reference to the test results of the completed scenarios, leading to low testing efficiency.

Compared with DOE-based methods, ADOE-based methods generate testing scenarios step by step following the idea of optimization. According to the testing results of previously generated testing scenarios, the parameters of new testing scenarios are adaptively designed to be more challenging for VUT. Currently, existing methods have applied various optimization algorithms in the safety testing for ADS. Crespo-Rodriguez et al. utilized a single-objective genetic algorithm (GA) to search for adversarial test scenarios \citep{cresporodriguez2024pafot}. Gladisch et al. used Systems Theoretic Process Analysis (STPA) to initiate a testing function scenario and then applied Bayesian Optimization (BO) to find hazardous combinations of test parameters and their values \citep{gladisch2019experiencea}. Our previous work used particle swarm optimization (PSO) \citep{feng2022multimodal} and monte-carlo tree search (MCTS) \citep{wu2024lambda} respectively to quickly find the multimodal distributed hazardous scenarios while guaranteeing their coverage. ADOE-based methods can quickly converge on the scenarios with high testing value and have shown great potential in efficiency, coverage and stability for safety testing of ADS. Therefore, ADOE-based methods are chosen as the basis for the evaluation stage in this paper.

In ADOE-based methods, another key issue that have garnered significant scholarly attention is when the testing should stop. One category of methods intuitively defined an upper limit on number \citep{zhang2024accelerated} or duration \citep{hellwig2019benchmarking} that an algorithm can be run. However, with these methods based on expert experience, it is difficult to accurately set the value of the stopping condition. Another category of methods defined the stopping condition from the perspective of algorithmic convergence, e.g., by finding a predefined optimal solution \citep{ravber2022maximum} or by using the convergence of a surrogate model that additionally trained using historical sampling records as the stopping criterion \citep{sun2022adaptive}. This type of methods can adaptively stop the sampling. However, they hold the issue of converging to local optima when there are multiple optimal solutions in ADS testing. To alleviate this effect, in this paper, we propose a more comprehensive approach to partitioning the training dataset for an observation model, whose prediction performance will be used as the stopping criterion.

\subsection{Safety Evaluation Methods for Autonomous Driving Systems}

With the testing results in hand, existing methods achieved safety evaluation for ADS either based on points (namely the testing results of concrete scenarios) or areas (namely hazardous domains) under a logical scenario space.

Points-based methods directly calculate the number or percentage of the hazardous concrete scenarios, in which the VUT does not fulfill the passing condition (a collision occurs or the safety metric exceeds a certain threshold), to represent the safety performance of VUT. Bussler et al. used an evolutionary algorithm to identify hazardous scenarios while the number of the identified scenarios was used to evaluate VUT \citep{bussler2020application}. Feng et al. regarded the accident rate as the evaluation metrics, where an accident was identified when the relative distance between the VUT and a virtual background vehicle was zero or less than zero under an augmented reality (AR) testing platform \citep{feng2020safety}. Ding et al. proposed a flow-based multimodal hazardous scenario generator and used collision rate as evaluation indicators \citep{ding2021multimodal}. ISO 34502 used the ratio of passing the scenario under test as a measure of safety evaluation \citep{iso34502}. Point-based methods can evaluate the safety level of VUT in logical scenarios to some extent, but they ignore the correlation of key parameters in the scenarios, which makes them difficult to meet the safety verification needs of high-level ADS.

Areas-based methods achieve safety evaluation by identifying hazardous domains, the number and distribution of which represent the safety performance of a VUT. Generally, hazardous domains can be obtained with classification or clustering methods. For instance, based on the sampled hazardous scenarios after testing, Mullins et al. used MeanShift, an unsupervised clustering method, to identify hazardous domains while getting their boundaries \citep{mullins2018adaptivea}. Similarly, Batshch et al. applied Gaussian Process Classification to identify the performance boundary through testing results obtained from Monte Carlo sampling and Latin Hypercube sampling \citep{batsch2019performancea}. Apart from the classification- and clustering-based methods, 
Zhang et al. introduced the gravitational field formed by the most hazardous scenarios identified during testing. By integrating this with the internal probability distribution of the logical scenario parameter space derived from naturalistic driving trajectory (NDT), the safety performance of the VUT at any point within the logical scenario could be calculated, which enabled the determination of hazardous domains \citep{zhang2023safety}.
However, most of the existing literature obtained hazardous domains with irregular boundaries, which hindered further quantitative evaluation of the safety of VUT, such as identifying the exact extent and calculating the actual size of the hazardous domains, especially in high-dimensional parameter space. To circumvent this restraint, this paper uses hyper-cuboid to approximate hazardous domains to obtain regular boundaries, which enables quantitative safety evaluation.

\section{Method} \label{method}

\subsection{Framework}

\begin{figure*}[t] 
      \centering
      \includegraphics[width=18cm]{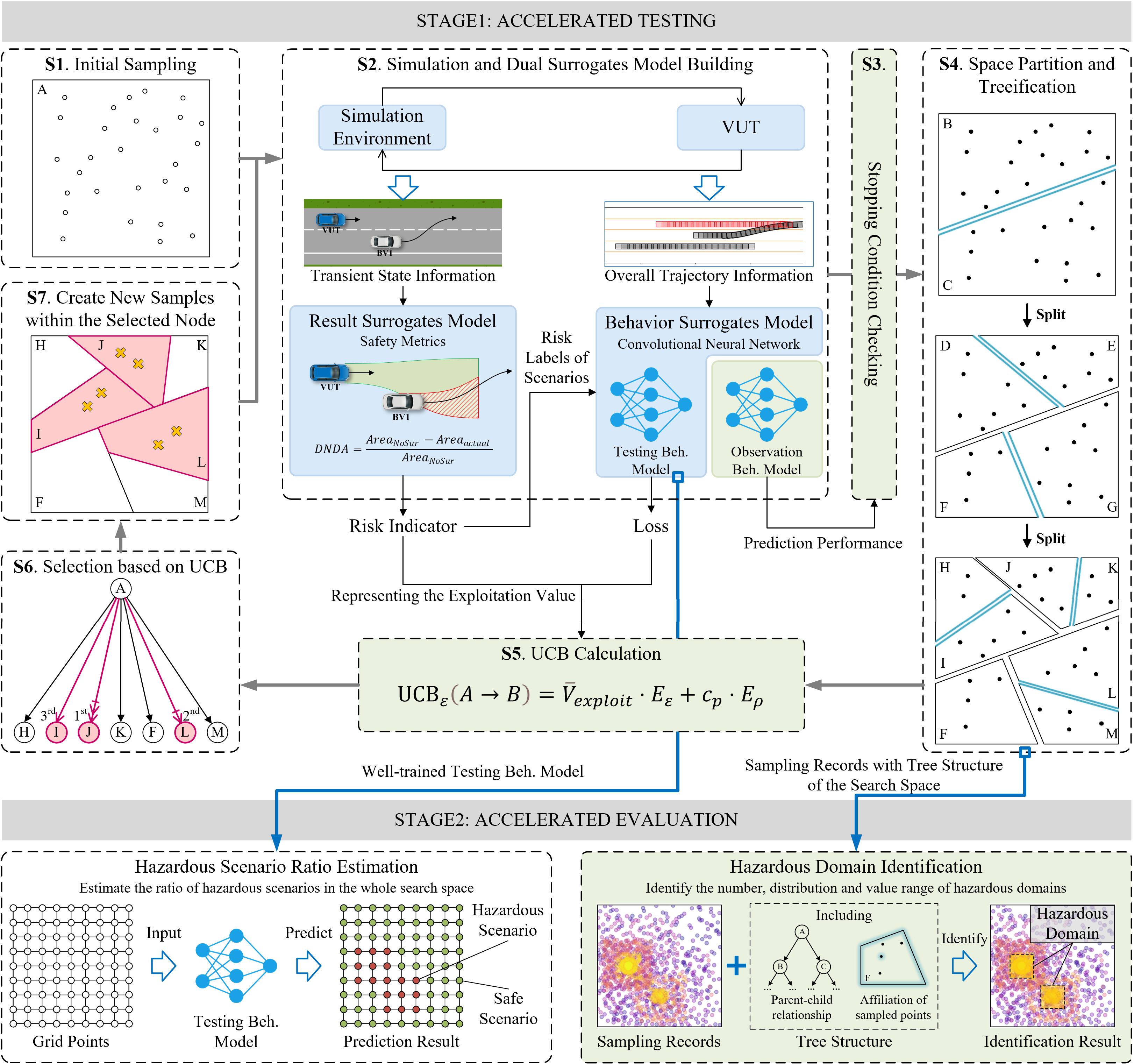}
      \caption{Framework of the Proposed Integrated Accelerated Testing and Evaluation Method (ITEM).}
      \label{framework}
\end{figure*}

The framework of the proposed Integrated Accelerated Testing and Evaluation Method (ITEM) is shown in Fig. \ref{framework}, which can be divided into accelerated testing stage and accelerated evaluation stage. The above two stages are integrated together to ensure that testing information is utilized as much as possible. The green parts highlight the improvements in this paper compared with our previous work \citep{wu2024accelerated, wu2024lambda}.

As depicted in Fig. \ref{framework}, in the accelerated testing stage, an initial random sampling is conducted at first. After that, concrete testing scenarios are generated based on the sampling results, which are then inputted to the in-loop testing. Next, the dual surrogates model is constructed, in which both the transient state information during testing and the overall trajectory information after testing are utilized. More in detail, during the testing, a certain safety metric is calculated using transient state information such as positions and velocities of ego and background vehicles to construct the result surrogate model (in this paper, we apply DNDA \citep{wu2022risk} as the safety metric). And after the testing, with a bird's eye view (BEV) recording the overall trajectory information and a risk label calculated by the safety metric as input, a CNN is used to construct the behavior surrogate model. It should be noted that an additional observation behavior surrogate model is trained in this paper, whose prediction performance will be used as a reference for the stopping condition. The settings of the observation behavior surrogate model and the construction of the stopping condition will be detailed in Sec. \ref{stopping_condition}. 

After checking the stopping condition, the entire logical space is recursively partitioned into good and bad subspaces based on risk values, and a tree structure is then constructed. Then, Upper Confidence Bound (UCB) of each node is calculated with the improved calculation method. Based on the UCB result, promising subspaces with a high probability of containing hazardous scenarios are chosen to generate new sampling points. The above process continues until the stopping condition is reached.

Once the accelerated testing stage is finished, the sampling records, as well as the well-trained testing behavior surrogate model and the tree structure of search space are transferred to the accelerated evaluation stage all together, where both points-based and areas-based methods are conducted for quantitative safety evaluation. To be specific, for points-based safety evaluation method, the testing behavior surrogate model is utilized as a proxy to quickly predict and evaluate the safety performance of the VUT in concrete scenarios that have not been tested before. Through the utilization of this surrogate model, the risk levels of each scenario obtained via fine grid sampling are predicted. Consequently, an estimation of the ratio of hazardous scenarios can be derived. At the same time, for areas-based safety evaluation method, based on the parent-child relationship of each subspace stemmed from the tree structure, hazardous scenarios in the sibling subspaces can be merged into the same hazardous domain, thus achieving hazardous domain identification. Detailed information can be found in Sec. \ref{evaluation}.

\subsection{Improvements on Accelerated Testing Stage}

Compared with our previous work \citep{wu2024accelerated, wu2024lambda}, two major improvements are proposed in this paper: 1) An improved UCB calculation method that emphasizing the value of boundary subspaces, 2) The stopping condition based on the prediction performance of an additionally trained observation model.

\subsubsection{Improved UCB Calculation Method}\label{improve_UCB}

As the basis for subspace selection, the computation result of UCB directly determines the direction and result of the searching of hazardous scenarios, and thus has been deeply investigated in our previous work. In \cite{wu2024lambda}, we have introduced density of subspaces into the calculation of UCB to overcome the sampling bias phenomenon. And in \citep{wu2024accelerated}, we have integrated UCB with the loss of CNN to choose the promising subspaces more accurately. In this paper, in order to serve the purpose of hazardous domain identification, it is necessary to find hazardous scenarios located at the boundary of the hazardous domain as many as possible. To this end, we first define boundary subspace as the subspace that contains both hazardous and safe scenarios, as shown in the following equation.

\begin{equation}
\label{eq1}
\begin{aligned}
G = G_b,\ if \ \text{max}(f(\boldsymbol{x}_i)) > f_b \ and \ \text{min}(f(\boldsymbol{x}_i)) < f_b, \\ i\in \{1,2,...,t\}
\end{aligned}
\end{equation}
where $G$ represents a subspace and $G_b$ is the identified boundary subspace. $\boldsymbol{x}_i$ is the $i^{th}$ sampled point belonging to $G$ that represents a concrete scenario. $f(\cdot)$ is a certain safety metric. And $f_b$ is the threshold to determine whether the scenario $\boldsymbol{x}_i$ is hazardous or not.

By attributing additional value to the identified boundary subspaces, the algorithm tends to select these boundary subspaces for searching, thereby achieving the aim of finding as many hazardous scenarios as possible at the boundary of the hazardous domains. The additional value of each boundary subspace $V_{boundary}$ can be calculated as:
\begin{equation}\label{eq2}
\begin{aligned}
V_{boundary} = \text{AVG}\Bigg[ \sqrt{\text{sin}\left(\frac{\{\text{min}[f(\boldsymbol{x}_{i,above})]-f_b\}\cdot\pi}{2(f_{up}-f_b)}\right)}, \\  \sqrt{\text{sin}\left( \frac{\{f_b-\text{max}[f(\boldsymbol{x}_{i,below})]\}\cdot\pi}{2(f_{b}-f_{low})} \right)} \Bigg]
\end{aligned}
\end{equation}
where $f_{up}$ and $f_{low}$ are the upper and lower bounds of the safety metric value. $\boldsymbol{x}_{i,above}$ and $\boldsymbol{x}_{i,below}$ represent all the points within the boundary subspace that above and below the threshold $f_b$. The formula first finds the closest scenarios to the threshold in $\boldsymbol{x}_{i,above}$ and $\boldsymbol{x}_{i,below}$, and then calculates the mean value of the distance of these two scenarios from the threshold to measure the degree of exploration of the hazardous domain boundary within the boundary subspace. A larger value indicates that the hazardous domain boundary is under-explored within the boundary subspace and more sampling is needed.

Additionally, in order to avoid the algorithm falling into a local optimum at the beginning due to over-sampling in the boundary subspaces, this paper introduces the idea of Random Dropout in neural network training. When the value of the boundary subspace is computed, this value is randomly deactivated with a probability $P_{drop}$, thus preventing over-sampling in certain boundary subspace. More in detail, the probability $P_{drop}$ in this paper is set to an adaptive value that decreases with the increase in the number of sampled points. By doing so, it can avoid the algorithm falling into local optima in the early sampling stage, while ensuring that the algorithm focuses on the boundary subspace in the late sampling stage, which is conducive to accurately identifying the boundary of the hazardous domains. The probability $P_{drop}$ is defines as:
\begin{equation}\label{eq3}
P_{drop}=\begin{cases}
    1-1/k\times N_{sample}, ~\text{if}\ N_{sample}< k \\
    0, ~~~~~~~~~~~~~~~~~~~~~~~~~~~~~\text{if}\ N_{sample} \ge k 
\end{cases}
\end{equation}
where $N_{sample}$ is the number of sampled points and $k$ is a tunable hyperparameter, which is determined by the dimension of the search space and the sampling budget.

Finally, the UCB calculation method that considers the subspace density, the loss of the CNN, and the value of the boundary subspace is shown in Eq. \ref{eq4}.
\begin{equation}\label{eq4}
\begin{aligned}
UCB_\epsilon(A\rightarrow B) = \bar{V}_{exploit} \cdot E_\epsilon + c_p \cdot E_\rho
\end{aligned}
\end{equation}

As shown in Eq. \ref{eq4}, the UCB from parent space $A$ to child space $B$ consists of two main terms. The former represents the exploitation value of subspace $B$, expressed as the product of the subspace value $\bar{V}_{exploit}$ the loss $E_\epsilon$ output by the result and behavior surrogates model on that subspace, respectively. The latter represents the exploration value of subspace $B$, expressed as the ratio of the density of sampled points in the parent space to the subspace (as shown in Eq. \ref{eq7}). $c_p$ is a tunable hyperparameter used to balance exploitation and exploration. 

Based on Eq. \ref{eq4}, three cases will be prioritized by the algorithm: 1) a high value $\bar{V}_{exploit}$ of the subspace, which indicates that the subspace contains numerous hazardous scenarios and is thus worthy of further exploitation, 2) a high loss $E_\epsilon$, which represents that the sampling in the subspace are not distinctly featured and the surrogate model is difficult to learn, and it is necessary to explore more to reduce the difficulty of learning, and 3) a high $E_\rho$, which represents that the subspace is not adequately explored, and it is necessary to explore more to ensure the coverage in the further search.

Further, in order to avoid a large difference between the two terms in the UCB which can lead to a failure of the balance between exploitation and exploration, it is necessary to ensure that the value domains of each term are in the same order of magnitude. In general, large values of the exploration term can cause the algorithm to converge slowly but this is acceptable to ensure coverage. However, large values of the exploitation term can cause the algorithm to fall into local optima, so it is necessary to limit the value of the exploitation term. In this paper, we first use a normalization function $N(x)$ to normalize the original values, followed by a convex function $G(x)$ to amplify the values of the nodes with low exploitation values so as to avoid falling into local optima. $N(x)$ and $G(x)$ can be expressed as:

\begin{equation}\label{eq5}
\begin{aligned}
N(x) = \frac{x-X_{low}}{\text{max}(x_{all})-X_{low}}
\end{aligned}
\end{equation}

\begin{equation}\label{eq6}
G(x)_=\begin{cases}
    \frac{1}{1-log_{10}^x}, ~x>0 \\
    0, ~~~~~~~~~~x=0 
\end{cases}
\end{equation}

In Eq. \ref{eq5}, $X_{low}$ is the lower bound of the value domain of $x$. Ultimately, the terms of Eq. \ref{eq4} are calculated as follows:

\begin{equation}\label{eq7}
\begin{aligned}
&\bar{V}_{exploit} = G[N(V_{exploit}+V_{boundary})] \\
&E_\epsilon = G[N(\bar{\epsilon}_B/\bar{\epsilon}_A)] \\
&E_\rho = \text{ln}(\bar{\rho}_A/\bar{\rho}_B)
\end{aligned}
\end{equation}
where $V_{exploit}$ is calculated by the safety metric $f(\cdot)$ and $V_{boundary}$ is the additional value of boundary subspace. $\bar{\epsilon}_A$ and $\bar{\epsilon}_B$ are the average loss of the behavior surrogate model in parent node $A$ and child node $B$. $\bar{\rho}_A$ and $\bar{\rho}_B$ are the average density of parent node $A$ and child node $B$. The calculations of $V_{exploit}$, $\bar{\epsilon}_{Node}$ and $\bar{\rho}_{Node}$ are essentially the same as in \citep{wu2024accelerated} and \citep{wu2024lambda}, but with minor modifications. The specific calculation methods of the above parameters can be found in Appendix \ref{appendix}.

\subsubsection{Stopping Condition} \label{stopping_condition}

In this paper, we define the stopping condition based on the prediction performance of the behavior surrogate model. If the model is able to predict accurately on the test set after training with the recorded sampled points, the search is sufficiently adequate to stop, otherwise the search should be continued. However, since all the recorded sampled points have been used to train the behavior surrogate model, there are no extra sampled points to test the prediction accuracy. 

To address this issue, in addition to the testing behavior surrogate model (hereinafter called the testing beh. model) that is trained using all of the sampled points, we construct in parallel an observation behavior surrogate model (hereinafter called the observation beh. model) that is trained using only a portion of the sampled points, with the remaining sampled points being used as the test set. By doing this, we actually construct a lite version of the behavior surrogate model. Existing studies have demonstrated that the predictive performance of the model improves as the number of samples in the training set increases \citep{sun2017revisiting} . Therefore, we can use the predictive performance of the observation beh. model to determine if the testing beh. model is adequately trained without affecting the search on the mainstream. 

\begin{figure}[b] 
      \centering
      \includegraphics[width=8cm]{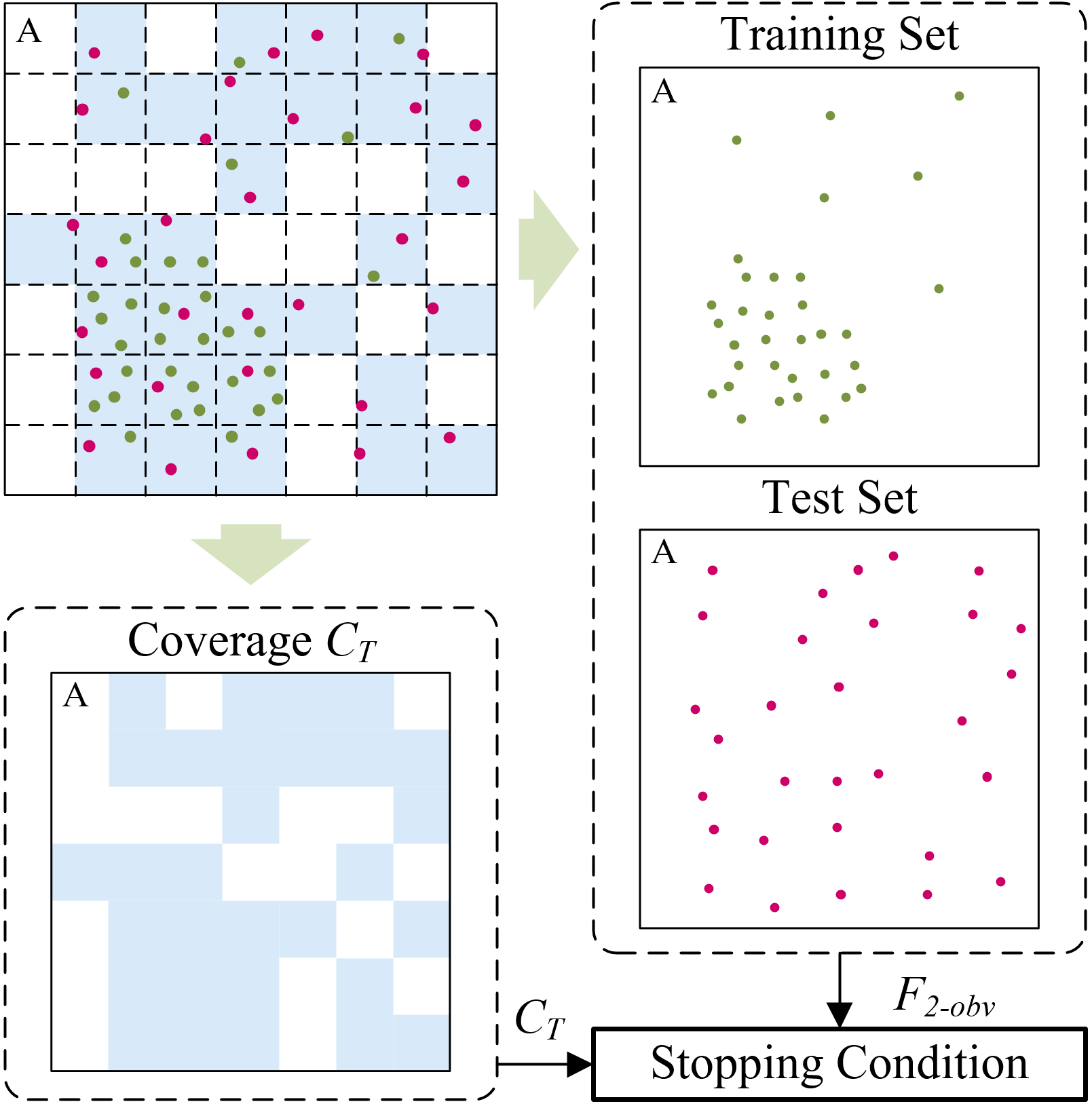}
      \caption{The stopping condition construction process.}
      \label{stop_condi}
\end{figure}

In order to be able to comprehensively measure the predictive performance of the observation beh. model, a representative test set covering the entire search space is required. However, since the purpose of the search is to find hazardous scenarios, the percentage of hazardous scenarios in the recorded sampled points is higher than the actual situation, so how to divide the test set is a key issue. This paper introduces the idea of spatial stratification in Latin Hypercube Sampling to form an unbiased test set. The stopping condition construction process is shown in Fig. \ref{stop_condi}.

As depicted in Fig. \ref{stop_condi}, the entire search space is first divided into equal grids (or cubes/hypercubes). After that, for each grid subspace containing historical sampled points, one of the sampled points is prioritized to be divided into the test set and the rest of the sampled points are divided into the training set. A clear example is illustrated in Fig. \ref{stop_condi}, where the test set evenly covers the entire search space. With the training set and the test set obtained, the observation beh. model is trained and then evaluated by Confusion Matrix Analysis, where we use $F_2-Score$ (denoted as $F_{2-obv}$) as metric to emphasize more on Recall. Concurrently, after the search space is divided, a quantitative test set coverage metric $C_T$ can be obtained from the ratio of the number of grids occupied by sampled points to the number of all grids. Finally, based on $F_{2-obv}$ and $C_T$, the stopping condition is established: The search will stop if both of the following conditions are met:

\begin{itemize}
    \item The test set must cover 80\% of the search space ($C_T \ge 80\%$) to ensure representative test results.
    \item The predictive performance metric $F_{2-obv}$ must be greater than the threshold $F_s$ ($F_{2-obv} \ge F_s$).
\end{itemize}

\subsection{Improvements on Accelerated Evaluation Stage} \label{evaluation}

As illustrated in Fig. \ref{framework}, both points-based and areas-based evaluation method are included in our framework. Since the points-based approach has been well described in \citep{wu2024accelerated}, this paper focuses on the areas-based approach to identify the hazardous domains.

Different from other studies that use only historical sampled points information, the hazardous domain identification method proposed in this paper also takes the tree structure into account. More in detail, two aspects of information are included in the tree structure transferred from the accelerated testing stage: 1) The affiliation of each historical sampled point with the subspaces, 2) The parent-child relationship between subspaces. 

For ease of interpretation, a schematic diagram is demonstrated in Fig. \ref{HD_iden}, where all the subspaces are partitioned in the way shown in Fig. \ref{framework}-S4, and assuming that the real distribution of the hazardous domains is as shown in the scatter plot in Fig. \ref{framework}. The identification method can be divided into four steps as follows:

\begin{figure}[t] 
      \centering
      \includegraphics[width=8cm]{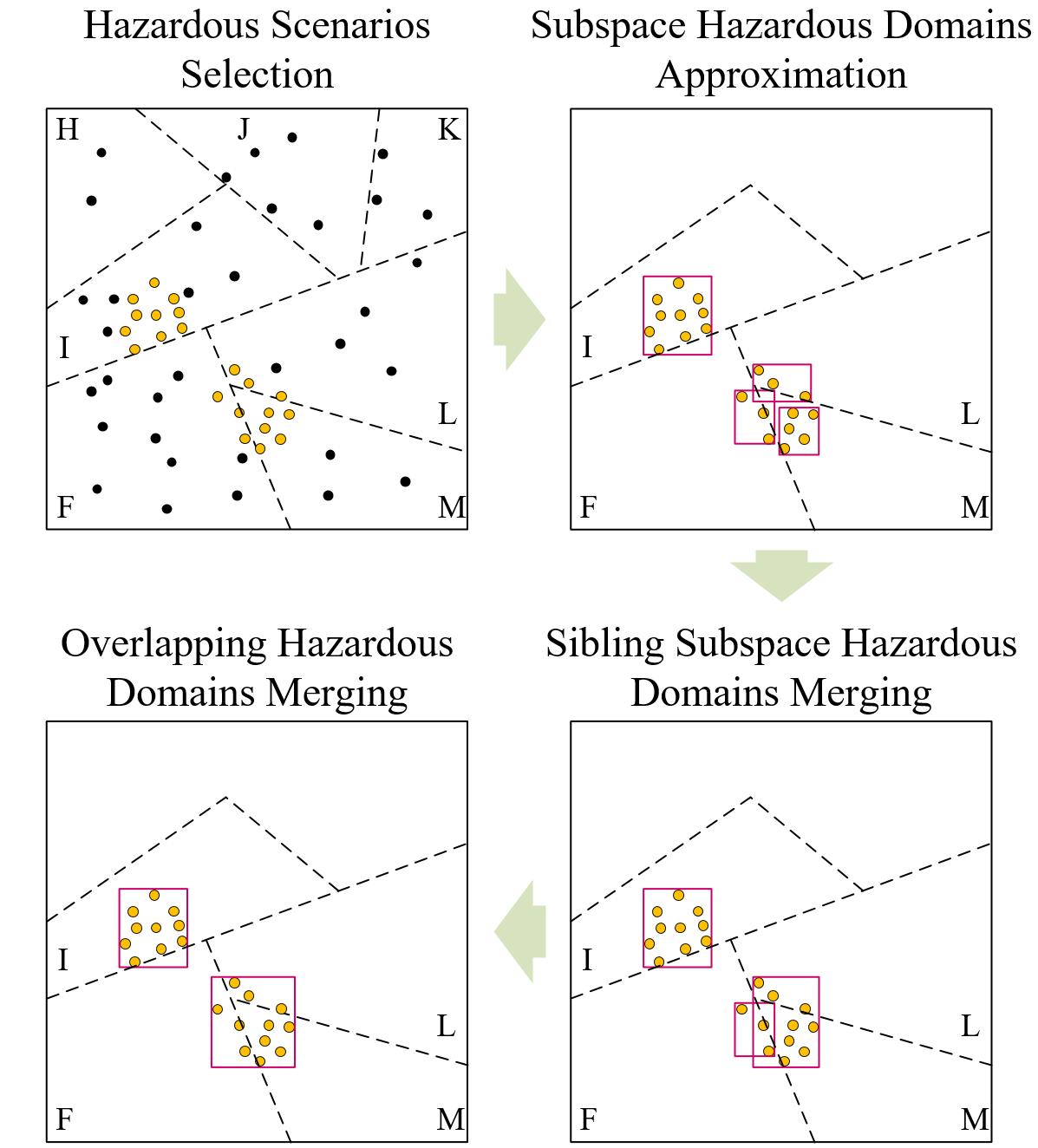}
      \caption{Hazardous domain identification method.}
      \label{HD_iden}
\end{figure} 

1) \textit{Hazardous Scenarios Selection}: All hazardous scenarios with safety metric values $f(\boldsymbol{x}_i)$ exceeding the threshold $f_b$ are selected first, as illustrated by the yellow dots in Fig. \ref{HD_iden}. After that, only subspaces containing hazardous scenarios are retained, while non-hazardous scenarios in these subspaces are also excluded.

2) \textit{Subspace Hazardous Domains Approximation}: Hyper-cuboids are used in this step to approximate the hazardous domain in each retained subspace, which are represented by the red boxes in Fig. \ref{HD_iden}. To be specific, the upper and lower bound of a hyper-cuboid in each dimension can be determined as:
\begin{equation}\label{eq8}
\begin{cases}
    B_{lower,j} = \text{min}(\boldsymbol(x)_{i,j}) \\
    B_{upper,j} = \text{max}(\boldsymbol(x)_{i,j})
\end{cases}
\end{equation}
where $i\in \{1,2,...,T\}$ represents the $i^{th}$ hazardous scenario in a certain subspace that has totally $T$ hazardous scenarios. $j\in \{1,2,...,dim\}$ represents the $j^{th}$ dimension that is being calculated in the $dim$-dimensional search space.

3) \textit{Sibling Subspace Hazardous Domains Merging}: With the parent-child relationship between subspaces known, hazardous domains belonging to a same parent subspace are merged. As shown in Fig. \ref{HD_iden}, since both the subspace $M$ and the subspace $L$ are partitioned from the subspace $G$, they are sibling subspaces and their hazardous domains are thus merged.

4) \textit{Overlapping Hazardous Domains Merging}: In case hazardous scenarios in the same region are identified into different hazardous domains, all the identified hazardous domains will be checked if they overlap with other hazardous domains and then merged. Note that only two hazardous domains overlapping in all dimensions will be merged. In Fig.\ref{HD_iden}, the hazardous domains in subspace $F$ and $G$ (parent subspace of $L$ and $M$) are merged because they overlap in all two dimensions, while the hazardous domain in subspace $I$ remains independent because it does not overlap with the merged hazardous domain. Finally, two hazardous domains are identified with their shapes represented by two rectangles.

\section{Experiments} \label{experiments}

\subsection{Synthetic Function and Practical Scenario for Testing}

In this paper, both synthetic function and practical scenario are considered. Meanwhile, the experiments cover cases from two to four dimensions, demonstrating the effectiveness of the method under both low and high dimensions.

\subsubsection{Two- and Four-dimensional Multimodal Gaussian Function}

This paper modifies the gaussian function, which only has one global optimum (i.e., one modality), into a multimodal gaussian function that can be defined as follows:

\begin{equation} \label{eq9}
\begin{aligned}
f(\boldsymbol{x})= \sum _{i = 1} ^ {dim} e^{-\frac{x_i^2}{2\sigma^2}}, \ \text{where} \ x_i = \Vert \boldsymbol{x} + M_{(i,:)} \Vert_2, \\ \text{for all} \  i = 1, 2, ..., dim
\end{aligned}
\end{equation}
in which,
\begin{equation} 
\label{eq10}
M =\begin{bmatrix}
	 1 & 0 & \cdots & 0 \\
      0 & 1 & \cdots & 0 \\
      \vdots & \vdots & \ddots & \vdots \\
      0 & 0 & \cdots &1
\end{bmatrix}_{dim} \times bias
\end{equation}
where $\boldsymbol{x}=(x_1,x_2,\cdots,x_{dim})$ represents a certain point in the search space. $dim$ is the dimension assigned to the function. $M_{(i,:)}$ represents the $i^{th}$ row of the matrix $M$. $bias$ and $\sigma$ are hyperparameters that determine the size and distribution of the hazardous domains of the function. Based on Eq. \ref{eq9}, the multimodal gaussian function with $dim$ dimensions has $dim$ modalities, which are respectively located in $x_1^*=(-bias,0,\cdots,0),x_2^*=(0,-bias,\cdots,0),\cdots,x_{dim}^*=(0,0,\cdots,-bias)$.

\begin{figure}[b] 
      \centering
      \includegraphics[width=8cm]{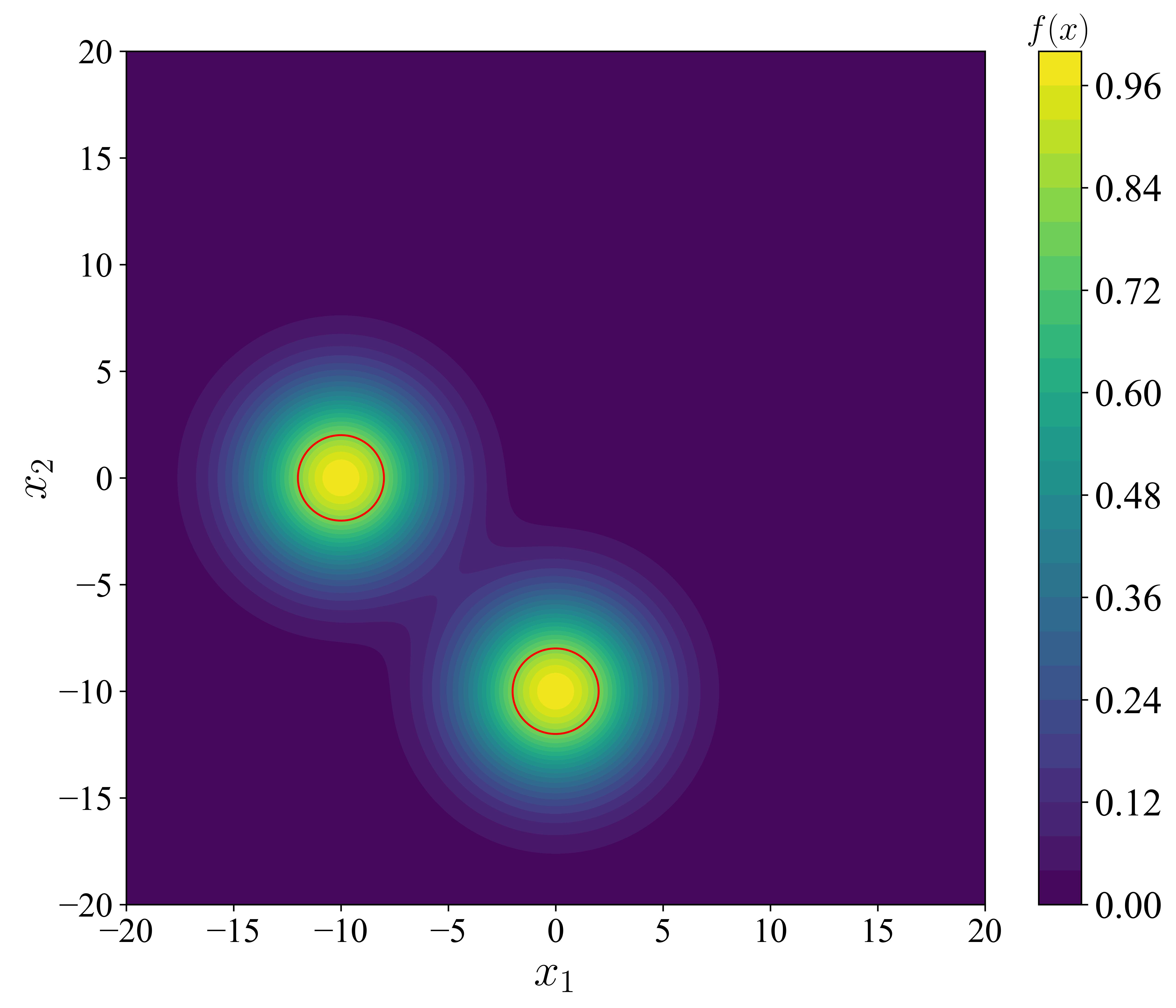}
      \caption{Illustration of 2-dimensional multi-modal Gaussian function.}
      \label{2dGauss_GT}
\end{figure} 

Specifically in this paper, the search space is built with a range of $x_i\in[-20,20]$ in each dimension, and the value domains of both functions are $[0,1]$. The hazardous threshold $f_b$ is chosen as 0.8 (i.e., points with a function value greater than 0.8 will be considered hazardous). Moreover, hyperparameters are set as $bias=10$ and $\sigma=3$ in both functions. For the two-dimensional function, an illustration is given in Fig. \ref{2dGauss_GT}, with red lines representing $f_b$. It can be seen that two modalities are located in $(0,-10)$ and $(-10,0)$ as expected. As for the four-dimensional function, by numerical calculation, it can be estimated that the percentage of the hazardous domains in the whole search space is $4.15\times10^{-5}$, which is consistent with the sparse distribution of hazardous scenarios in the real world, and at the same time challenging enough as a test function.

\subsubsection{Three-dimensional Cut-in Scenario}

\begin{figure}[h] 
      \centering
      \includegraphics[width=8cm]{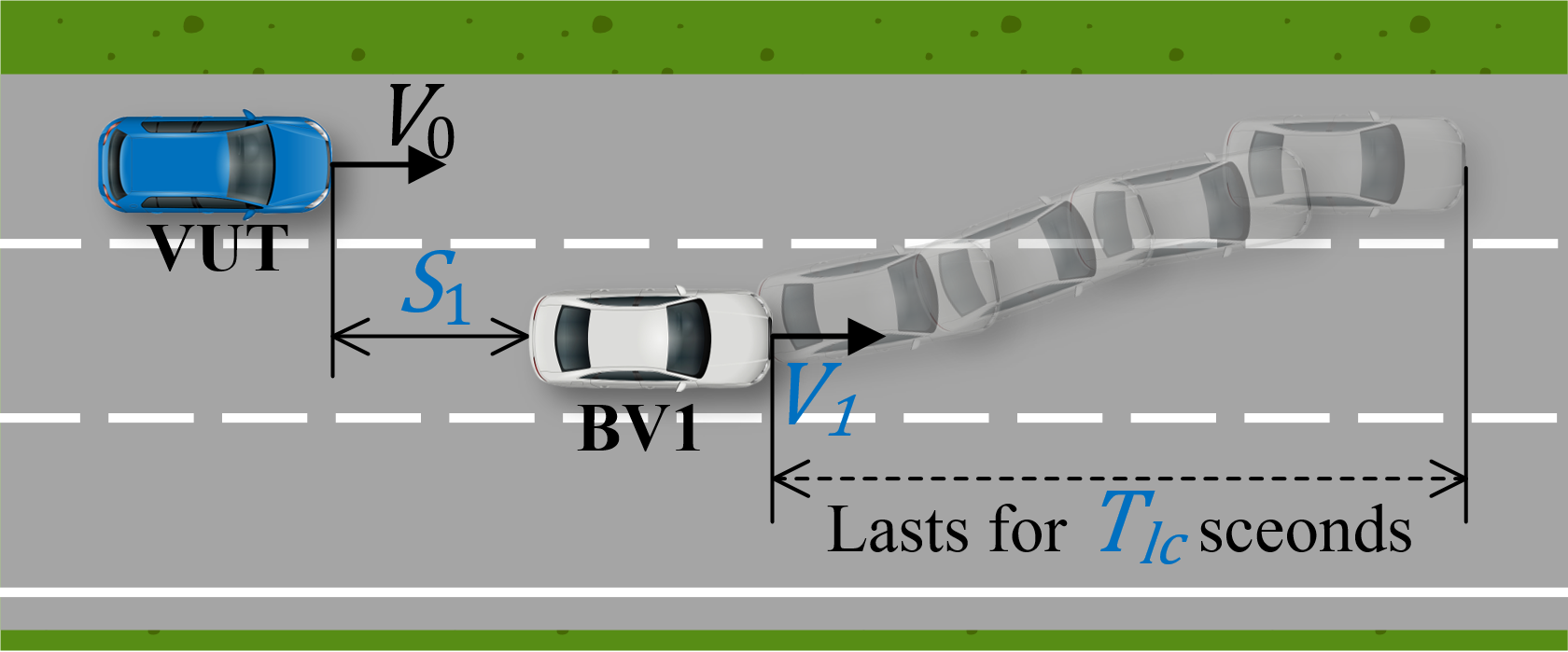}
      \caption{Three-dimensional logical testing scenario.}
      \label{3dScenario}
\end{figure}

Based on the simulation platform Virtual Test Drive (VTD) \citep{hexagon2025}, a logical testing scenario is conducted. As demonstrated in Fig. \ref{3dScenario}, the VUT, which is controlled by the built-in ADS in VTD, is driving in the left lane with an initial speed of $V_0$, ahead of which a background vehicle BV1 is driving in the middle lane at a distance $S_1$. At $t$ seconds after the start of the test, BV1 will perform a cut-in maneuver at the speed of $V_1$, and the whole cut-in process will last for $T_{lc}$ seconds. Detailed parameters of this logical testing scenario can be found in Table \ref{tab1}, where $S_1$, $V_1$ and $T_{lc}$ are parameters that construct the 3-dimensional search space, while $V_0$ and $t$ are fixed parameters. Additionally, we choose DNDA \citep{wu2022risk} as the safety metric. DNDA is a normalized risk indicator based on drivable area. The closer its value is to 1, the more hazardous the scenario is. And a collision occurs when its value is equal to 1.

\begin{table}[width=\linewidth,cols=4,pos=b]
\renewcommand{\arraystretch}{1.3}
\caption{Parameter Settings of the Logical Testing Scenario. }
\label{tab1}
\begin{center}
\begin{tabular}{ m{0.36\linewidth}  m{0.18\linewidth}<{\centering} m{0.18\linewidth} m{0.08\linewidth}}
\toprule
\textbf{Parameter} & \textbf{Denotation} & \textbf{Value/Range} & \textbf{Unit}\\
\midrule
Initial velocity of VUT & $V_0$  &  30 & $m/s$ \\

Initial distance between VUT and BV1 & $S_1$  &  $[30,110]$ & $m$ \\

Initial velocity of BV1 & $V_2$  &  $[20,30]$ & $m/s$ \\

Duration of lane change for BV1 & $T_{lc}$  &  $ [2,3] $  & $s$ \\

Starting moment of lane change for BV1 & $t$  &  3 & $s$\\
\bottomrule
\end{tabular}
\end{center}
\end{table}

\begin{figure}[t] 
      \centering
      \includegraphics[width=8cm]{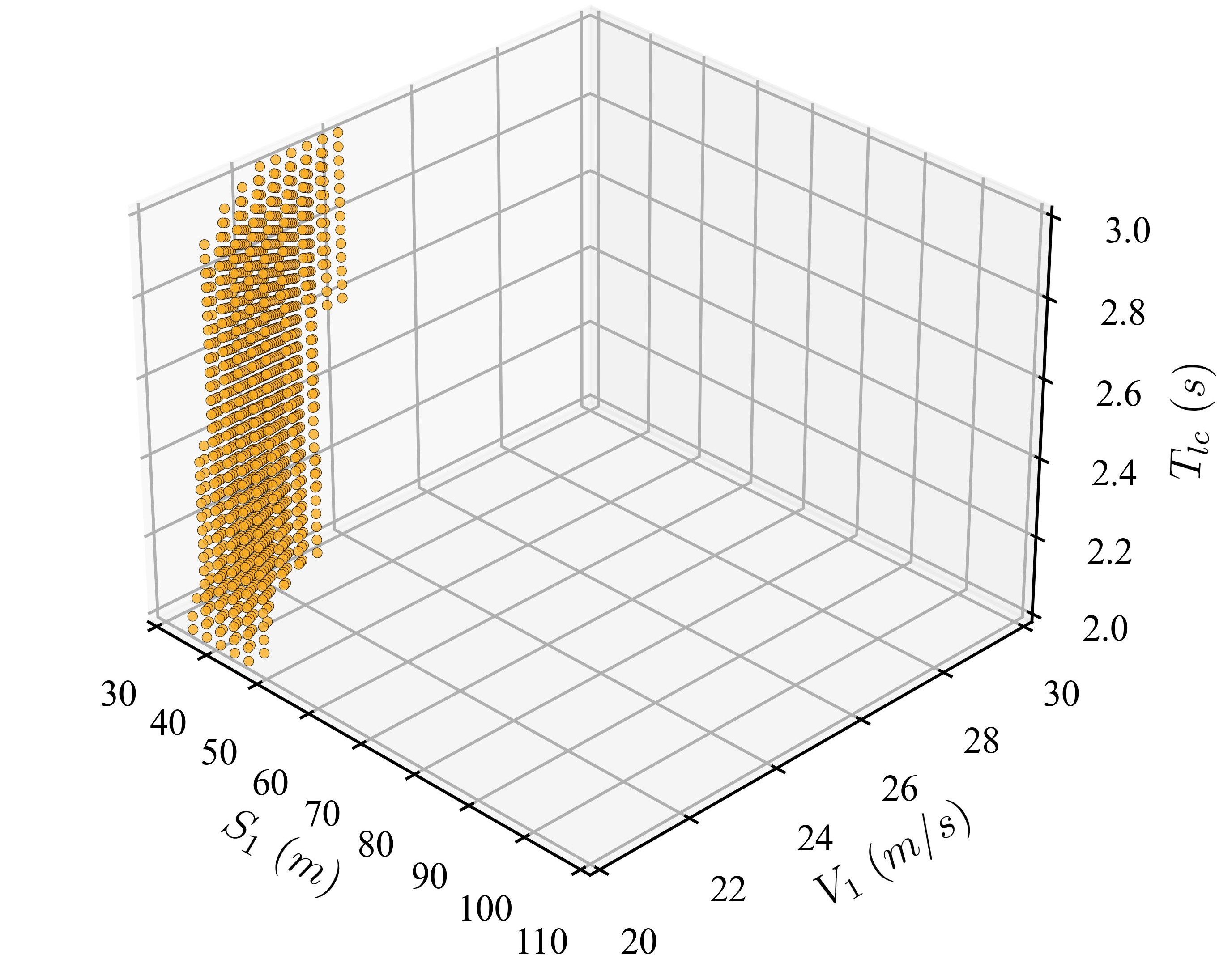}
      \caption{Hazardous scenarios distribution in the logical testing scenario.}
      \label{3dScenario_GT}
\end{figure}

Unlike synthetic functions where the distribution of hazardous points can be quickly obtained through numerical computation, the distribution of hazardous scenarios in the logical scenario space needs to be obtained through simulation. To this end, a $30\times30\times30$ grid testing is executed to obtain the ground truth distribution of hazardous scenarios, which can be used as a reference for subsequent evaluation. In this experiment, scenarios with a maximum DNDA value 
of more than 0.8 are considered hazardous. For ease of visualization, Fig. \ref{3dScenario_GT} only shows the hazardous scenarios.

\subsection{Algorithms to be Tested}
To validate the effectiveness of the improved method proposed in this paper as well as its superiority compared to other baseline methods, both the ablation experiment and the comparative experiment are conducted. The tested algorithms are detailed as follows.

1) \textit{Integrated accelerated Testing and Evaluation Method (ITEM)} : ITEM is the method proposed in this paper, which uses the improved UCB considering the value of boundary subspaces as described in Sec. \ref{improve_UCB} and makes full use of the testing information in the evaluation stage.

2) \textit{Integrated accelerated Testing and Evaluation Method with the Original UCB (ITEMoriUCB)} : Compared with ITEM, ITEM with the original UCB uses the original UCB calculation method in \cite{wu2024accelerated} , which doesn't take the boundary value into account.

3) \textit{Gaussian Distribution Method with Optimization Sampling (GDMos))} : GDM is an evaluation method proposed in \cite{zhu2022hazardous}, which uses gaussian distribution to cluster the hazardous scenarios and then obtains the hazardous domains. GDM is originally input with sampled points obtained by the optimization search method in \cite{zhu2022hazardous}. Since this evaluation method is independent of the testing stage, for the fairness of the comparison, this paper uses the sampling results of ITEM in the accelerated testing stage as the input to GDM.

4) \textit{Gaussian Distribution Method with Random Sampling (GDMrs)} :To further investigate the impact of different search algorithms in the testing stage on the evaluation stage, this paper also conducts random sampling and inputs the sampling results into the GDM. 

Detailed information of each algorithm is listed in Table \ref{tab2}. It should be noted that the input to the behavior surrogate model must be scenario trajectory information. Consequently, neither the behavior surrogate model nor the stopping condition is available on the two synthetic functions. Thus, for the 2- and 4-d multimodal gaussian functions, this paper only applies the result surrogate model in the accelerated testing stage and uses a predefined sampling budget as the stopping condition.

\begin{table}[width=10cm,cols=5,pos=t]
\renewcommand{\arraystretch}{1.3}
\caption{Experimental Settings of the Algorithms.}
\label{tab2}
\begin{center}
\begin{tabular}{m{0.85cm}  m{1.6cm} m{1.4cm} m{1.3cm} m{1.2cm}}
\toprule
\textbf{Test case} & \textbf{Algorithm} & \textbf{UCB Selection} & \textbf{Surrogate Model} & \textbf{Sampling Budget}\\
\midrule
\multirow{4}{0.85cm}{2-d function} & ITEM & Improved & \multirow{3}{1.3cm}{Single (only res)} & \multirow{4}{*}{900} \\ \cline{2-3}
~ & ITEMoriUCB & Original & ~ & ~ \\ \cline{2-3}
~ & GDMos & Improved & ~ & ~ \\ \cline{2-4}
~ & GDMrs & $\backslash$ & $\backslash$ & ~ \\ \hline

\multirow{4}{0.85cm}{3-d scenario} & ITEM & Improved & \multirow{3}{1.3cm}{Dual (res+beh)} & \multirow{4}{1.2cm}{Depend on the stopping condition} \\ \cline{2-3}
~ & ITEMoriUCB & Original & ~ & ~ \\ \cline{2-3}
~ & GDMos & Improved & ~ & ~ \\ \cline{2-4}
~ & GDMrs & $\backslash$ & $\backslash$ & ~ \\ \hline

\multirow{4}{0.85cm}{4-d function} & ITEM & Improved & \multirow{3}{1.3cm}{Single (only res)} & \multirow{4}{*}{30000} \\ \cline{2-3}
~ & ITEMoriUCB & Original & ~ & ~ \\ \cline{2-3}
~ & GDMos & Improved & ~ & ~ \\ \cline{2-4}
~ & GDMrs & $\backslash$ & $\backslash$ & ~ \\
\bottomrule
\end{tabular}
\begin{tablenotes}    
        \footnotesize
        * res: result surrogate model, beh: behavior surrogate model
    \end{tablenotes}
\end{center}
\end{table}

\subsection{Evaluation Metrics}

The main purpose of this paper is to make full use of the intermediate testing information into the safety evaluation, thus focusing more on the effect of improvements on the evaluation stage. Therefore, this paper proposes two indicators to quantify the evaluation performance, namely the accuracy of hazardous domain distribution identification (ADI) and the accuracy of hazardous domain percentage identification (API). For the evaluation metrics in the testing stage, this paper follows the $F_2-Score$ mentioned in \cite{wu2024lambda} (denoted as $F_{2-grid}$ in this paper).

1) $F_{2-grid}$ : $F_2-Score$ is a typical evaluation metric that widely used in machine learning. Compared with $F_{2-obv}$ which uses the test set divided from the sampling records as the actual values in confusion matrix, $F_{2-grid}$ uses the grid testing data in Fig. \ref{3dScenario_GT} to see if the algorithm correctly predicts the values at the grid points. Noted that calculating $F_{2-grid}$ takes more computational resources than calculating $F_{2-obv}$ because the predicted values for all grid points need to be calculated. Therefore, $F_{2-grid}$ is only used for the evaluation of different algorithms and not for the determination of the stopping conditions.

2) \textit{Accuracy of Hazardous Domain Percentage Identification (API)} : API is used to measure the percentage of ground truth hazardous domains covered by identified hazardous domains. Suppose there are $n$ hazardous domains in the $dim$-dimensional search space $A$, and the range of each ground truth hazardous domain $HD_{GT}^i$ in each dimension is $[a_d^i,b_d^i]$, where $i={1,2,\cdots,n}$, $d={1,2,\cdots,dim}$. Meanwhile, after the evaluation stage, totally $m$ hazardous domains $HD_{ID}^j$ are identified with the range $[p_d^j,q_d^j]$ on each dimension, where $j={1,2,\cdots,m}$, $d={1,2,\cdots,dim}$. Then API can be calculated as:

\begin{equation}\label{eq11}
\begin{aligned}
API = \frac{1}{2n}\sum^n_{i=1} \left( \frac{V^i_{Overlap}}{V^i_{GT}} + \frac{V^i_{Overlap}}{V^{sum \sim i}_{ID}} \right)
\end{aligned}
\end{equation}
where $V_{Overlap}^i$ represents the (hyper)volume of overlap of all identified hazardous domains with the $i^{th}$ ground truth hazardous domain, whose (hyper)volume is $V_{GT}^i$. $V^{sum \sim i}_{ID}$ is the sum of the (hyper)volumes of all identified hazardous domains that overleap with $HD_{GT}^i$. Algorithm \ref{alg1} shows the calculation method of $V_{Overlap}^i$. 

\begin{algorithm}[b]
\renewcommand{\algorithmicrequire}{\textbf{Input:}}
\renewcommand{\algorithmicensure}{\textbf{Output:}}
\caption{Calculation Method of $V_{Overlap}^i$ } 
\label{alg1} 
\begin{algorithmic}[1]
    \REQUIRE \ \\
    $[a_d^i, b_d^i]$, Upper and lower boundaries of $i^{th}$ ground truth hazardous domains in each dimension;  \\
    $[p_d^j, q_d^j]$, Upper and lower boundaries of $j^{th}$ identified hazardous domains in each dimension; \\
    $m$, \hspace{0.8cm} Number of identified hazardous domains;  \\
    $dim$, \hspace{0.48cm} Dimension of the search space.
    
    \ENSURE \ \\
    $V_{Overlap}^i$, Overlapping volume.
    
    \STATE Volume $\leftarrow$ 0
    \FOR{$j=1$ to $m$} 
        \STATE V $\leftarrow$ 0
        \FOR{$d=1$ to $dim$}
            \STATE Begin $\leftarrow$ max($a_d^i, p_d^j$)
            \STATE End $\leftarrow$ min($b_d^i, q_d^j$)
            \STATE Len $\leftarrow$ max(0, End-Begin)
            \STATE V $\leftarrow$ V $\times$ Len
        \ENDFOR
        \STATE Volume = Volume+V
    \ENDFOR
    \RETURN Volume
\end{algorithmic} 
\end{algorithm}

3) \textit{Accuracy of Hazardous Domain Distribution Identification (ADI)} : ADI is used to quantify the extent to which the spatial distribution of identified hazardous domains differs from the ground truth hazardous domains. Using the same notations, the distance between the centroids of two hazardous domains, $HD_{GT}^*$ and $HD_{ID}^*$, which are determined as overlapping in accordance with Algorithm \ref{alg1}, can be computed via Equation \ref{eq12}.

\begin{equation}\label{eq12}
\begin{aligned}
D^j_{center} = \left. \sqrt{\sum_{d=1}^{dim}\left[ \frac{(a_d^i+b_d^i)-(p_d^j+q_d^j)}{2}\right]^2} \right|_{i=*,j=*}
\end{aligned}
\end{equation}

Then the ADI of $HD_{ID}^*$ can be obtained.

\begin{equation}\label{eq13}
\begin{aligned}
ADI_j = 1 - \left. \frac{D^j_{center}}{D^i_{max}} \right|_{i=*,j=*}
\end{aligned}
\end{equation}
where $D^i_{max}$ is the distance from the centroid of $HD_{GT}^*$  to a certain vertex. Finally, the ADI of all the identified hazardous domains are calculated as shown in Eq. \ref{eq14}.

\begin{equation}\label{eq14}
\begin{aligned}
ADI = \frac{1}{n}\sum_{i=1}^n \frac{1}{l} \sum_{j=x}^{x+l} ADI_j
\end{aligned}
\end{equation}

Considering the case that a same ground truth hazardous domain may be identified as multiple hazardous domains, Eq. \ref{eq14} first calculates the average of the $ADI_j$ for a total of $l$ identified hazardous domains that all overlap with $HD_{GT}^i$, and then calculate the final $ADI$.

\subsection{Results and Analysis}

The results of the evaluation metrics for each algorithm are shown in the Table \ref{tab3}. All the evaluation metrics are calculated after the sampling budgets are run out. In the 3-d cut-in scenario experiment, the search is stopped at 2750 samples according to the stopping condition, which is detailed in Sec. \ref{sec_stop_condi}.

\begin{table*}[t]
\renewcommand{\arraystretch}{1.3}
\caption{Results of the Evaluation Metrics for Each Algorithm.}
\label{tab3}
\begin{center}
\begin{tabular*}{\linewidth}{m{2cm} m{1.1cm}<{\centering} m{1.1cm}<{\centering} m{1.1cm}<{\centering} m{0.2cm} m{1.1cm}<{\centering} m{1.1cm}<{\centering} m{1.1cm}<{\centering} m{0.2cm} m{1.1cm}<{\centering} m{1.1cm}<{\centering} m{1.1cm}<{\centering}}
    \toprule
     \multirow{2}{*}{\textbf{Algorithm}} & \multicolumn{3}{c}{\textbf{2-d function}} & ~ &\multicolumn{3}{c}{\textbf{3-d scenario}} & ~ & \multicolumn{3}{c}{\textbf{4-d function}} \\ \cline{2-4} \cline{6-8} \cline{10-12}
     ~ & $F_{2-grid}$ & API & ADI & ~ & $F_{2-grid}$ & API & ADI & ~ & $F_{2-grid}$ & API & ADI \\ 
    \midrule
    ITEM (ours) & 0.985 & \textbf{0.965} & \textbf{0.993} & ~ & \textbf{ 0.947} & \textbf{0.911} & \textbf{0.911} & ~ &  \textbf{0.962} & \textbf{0.845} & \textbf{0.980} \\
    ITEMoriUCB & \textbf{0.990} & 0.950 & 0.973 & ~ &  0.953 & 0.876 & 0.778 & ~ &  0.952 & 0.840 & 0.973 \\
    GDMos & 0.985 & 0.765 & 0.966 & ~ &  0.947 & 0.626 & 0.593 & ~ &  0.962 & 0.739 & 0.934 \\
    GDMrs & 0.829 & 0.719 & 0.938 & ~ &  0.855 & 0.629 & 0.642 & ~ &  0.029 & \text{NaN} & \text{NaN} \\
    \bottomrule
\end{tabular*}
\end{center}
\end{table*}

\subsubsection{The Efficiency of the Proposed ITEM}

As shown in Table \ref{tab3}, ITEM achieves the best performance compared with other baseline methods on both API and ADI metrics, showing the effectiveness and superiority of the proposed method in this paper. 

Compared to ITEMoriUCB that uses the original UCB calculation method, ITEM obtains higher API and ADI in the evaluation stage, as well as an approximately equal $F_{2-grid}$ in the testing stage, indicating that ITEM enhances the evaluation performance without affecting the testing efficiency. This part will be detailed in Sec. \ref{sec_UCB}. Compared to GDMos which uses the same sampling records from the testing stage, ITEM outperforms it by 20\%, 28.5\% and 10.6\% respectively for the three cases in terms of API, indicating that the identified hazardous domains of ITEM cover more of the ground truth hazardous domains and have fewer overestimations.

When it comes to the performance of ITEM on different test cases, it can be seen that in the two low-dimensional cases, both API and ADI of ITEM exceed 0.9, representing an accurate identification of the percentage and distribution of hazardous domains. While in the 4-dimensional case, ITEM obtains a high ADI of 0.980 but a relatively low API. By analyzing the detailed boundary data, we find that the identified hazardous domains achieve essentially more than 90\% coverage in each dimension of each modality, but since the hypervolume calculation in Eq.\ref{eq11} is multiplicative, the volume ratio will inevitably decrease as the dimensions get higher. Therefore, a slightly lower API is acceptable for high-dimensional cases.

\subsubsection{The Efficiency of the Improved UCB} \label{sec_UCB}

\begin{figure}[t]%
    \centering
    \subfigure[ITEMoriUCB]{
        \label{ITEMori}
        \includegraphics[width=.46\linewidth]{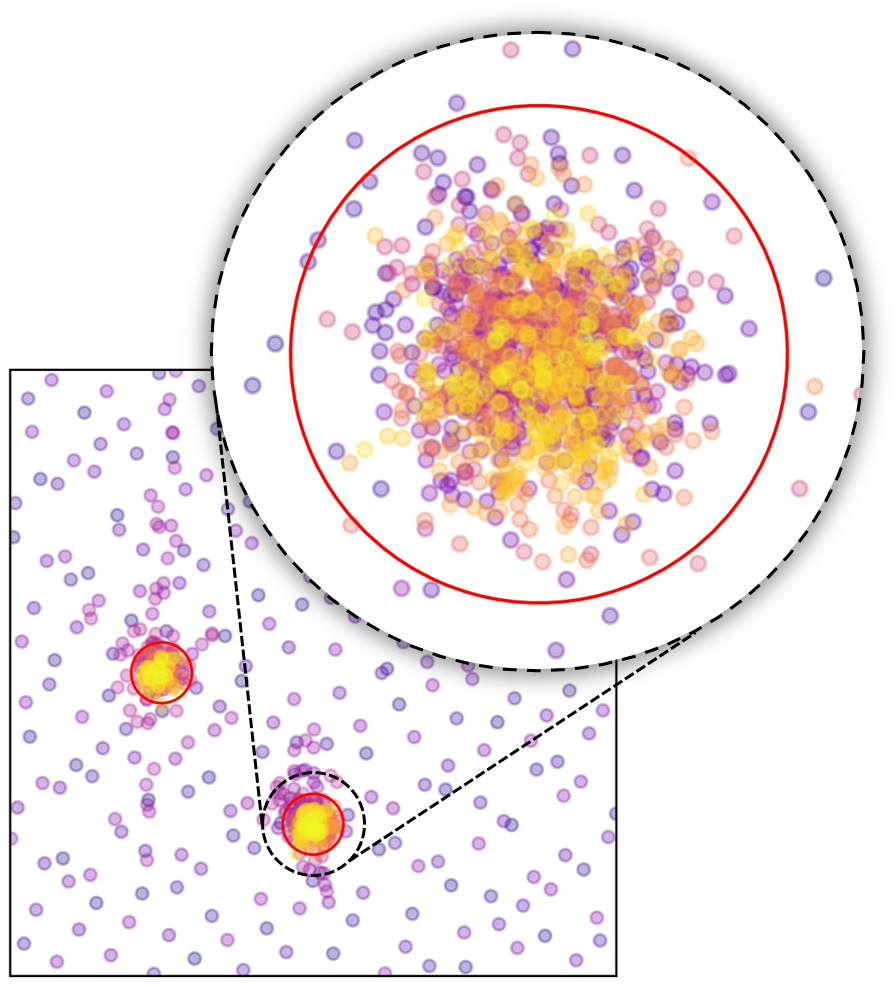}
        }
    \subfigure[ITEM]{
        \label{ITEM}
        \includegraphics[width=0.46\linewidth]{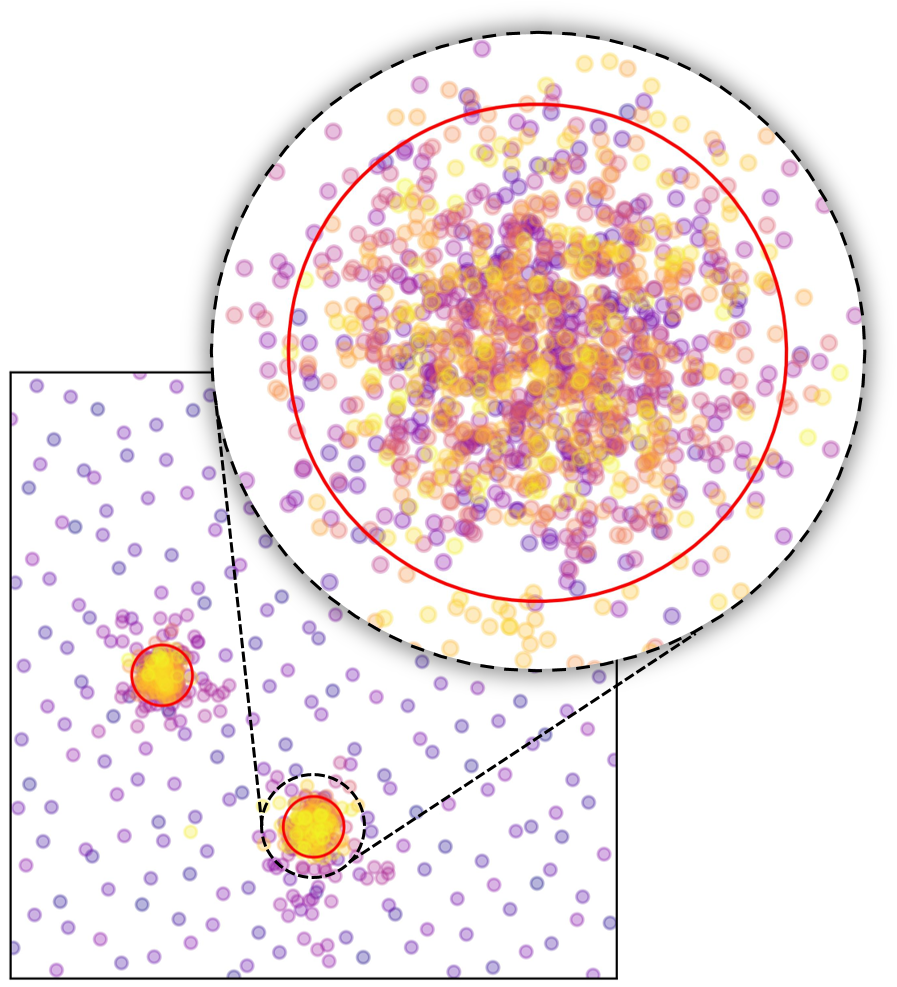}
        }
    \caption{Illustration of the sampling dynamics. Color from purple to golden represents the record being sampled at the beginning to the end of the optimization.
    }
    \label{illu_UCB}
\end{figure}

As discussed above, the use of the improved UCB does not diminish the efficiency of the algorithm in the testing stage (in some cases the search efficiency is even higher), while it significantly improves the accuracy of the hazardous domain identification in the evaluation stage. An illustration of the sampling dynamics of ITEMoriUCB and ITEM on the 2-d test case is shown in Fig. \ref{illu_UCB}. For the entire search space, both algorithms converge quickly to the two hazardous domains of the synthetic function after only a small amount of global sampling. However, when we concentrate on a certain modality, a clear difference between the two algorithms is observed. As shown in the magnified parts of Fig. \ref{illu_UCB}, ITEMoriUCB focuses only on the most hazardous regions and continues sampling on the center of the modality, whereas ITEM places more samples in the vicinity of the boundary of the hazardous domain, resulting in a full exploration of the hazardous domain boundary.

Note that even though there is a large difference between the two algorithms in the visualization results of the 2-d case, this difference is not obviously reflected in the computed API and ADI. We speculate that this is because the ground truth hazardous domains are circles, and even if ITEMoriUCB only focuses on sampling at modality centers, there will still be sporadic sampled points falling near the boundary of the hazardous domains, which improves the accuracy of hazardous domain identification. This phenomenon does not occur in the 3-d case, since its ground truth hazardous points show a rectangular distribution.

\subsubsection{The Impact of the Shape of the Ground Truth Hazardous Domains}

From the above analysis, we can find that the shape of the ground truth hazardous domain has a significant effect on the accuracy of hazardous domain identification. Furthermore, it can be found from Table \ref{tab3} that all the algorithms get a higher value of ADI on both synthetic functions, whereas the ADI results are lower in the cut-in scenario, except for ITEM. We argue that this is because the ground truth hazardous points are isotropically distributed in both synthetic functions, so that even if the sampling is focused only on the modality center, it can still guarantee the accuracy of the distribution identification. As a comparison, since the ground truth hazardous points in the 3-d case are anisotropically distributed, all the baseline algorithms focusing only on the center of modalities do not have a high ADI in this case, while the method proposed in this paper still obtains an ADI of more than 0.9.

In practice, in some relative studies \citep{sun2022adaptive, zhu2022hazardous}  and standards such as UN ECE R157 \citep{r157} , the distribution of the hazardous scenarios are always anisotropical and the ground truth hazardous domains vary in shape. Therefore, an algorithm that is independent of the shape of hazardous domains is needed to achieve accurate identification of hazardous domains in various scenarios. The results in Table \ref{tab3} show that the ITEM proposed in this paper is highly versatile and has great potential for hazardous domain identification in diverse logical scenarios.

\subsubsection{The Impact of Search Algorithms on the Evaluation Stage}

For GDM, in our experiment we respectively use the sampled points obtained from the optimization search and the random search as inputs for hazardous domains identification. From Table \ref{tab3}, it can be found that the identification results obtained based on the optimization sampled points are better than those obtained based on the randomly sampled points. And this gap is particularly noticeable in the 4-d case, where the $F_{2-grid}$ of GDMrs is only 0.029 because of the high dimensionality and the sparse distribution of hazardous points, resulting in almost no hazardous points available for hazardous domain identification. The above discussion illustrates that sampling efficiency in the testing stage has a significant impact on the evaluation stage, especially in a high-dimensional search space. Since the efficiency and coverage of the optimization algorithm used in our method have been well validated in \cite{wu2024lambda}, ITEM can be provided with a comprehensive input in the evaluation stage to ensure an accurate identification of the hazardous domains.

\subsubsection{Analysis of the Application of the Identified Hazardous Domains}

\begin{figure}[b] 
      \centering
      \includegraphics[width=8cm]{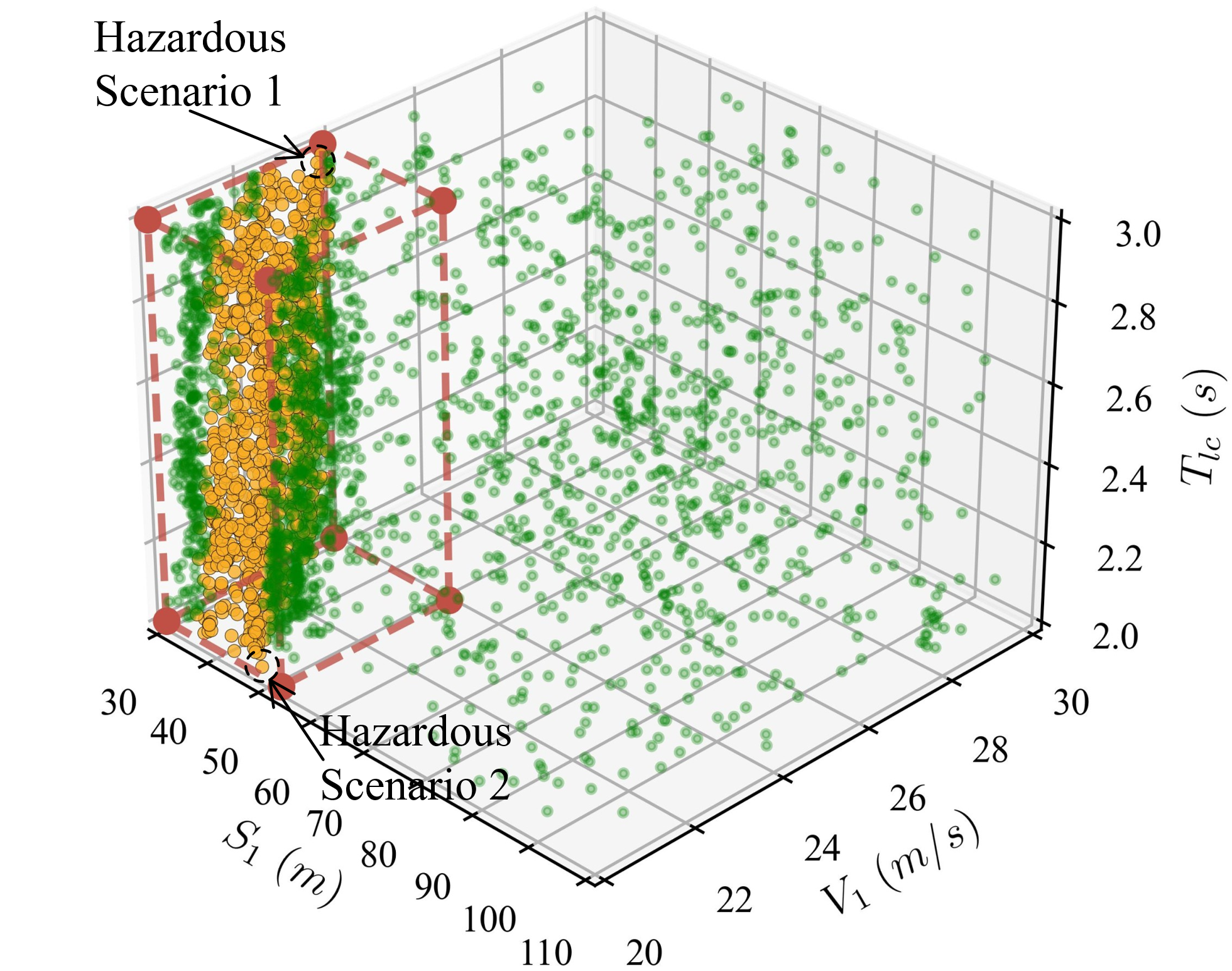}
      \caption{Testing and evaluation results of ITEM on the 3-d cut-in scenario.}
      \label{3dScenario_res}
\end{figure}

Based on experimental results of the 3-d cut-in scenario, this section delves into the real-world implications of an identified hazardous domain within a practical context. Moreover, this section conducts a further examination of how the results of hazardous domain identification can be applied to safety evaluation. The testing and evaluation results of ITEM on the 3-d cut-in scenario is shown in Fig. \ref{3dScenario_res}, where the yellow dots represent the searched hazardous scenarios (these scenarios have a DNDA > 0.8), while the green dots represent the searched safe scenarios. The red dashed rectangular box in Fig. \ref{3dScenario_res} is the hazardous domain identified by ITEM.

\begin{figure}[b]%
    \centering
    \subfigure[Hazardous Scenario 1. $(S_1=31.48m,V_1=23.53m/s,T_{lc}=2.96s)$]{
        \label{sce2722}
        \includegraphics[width=\linewidth]{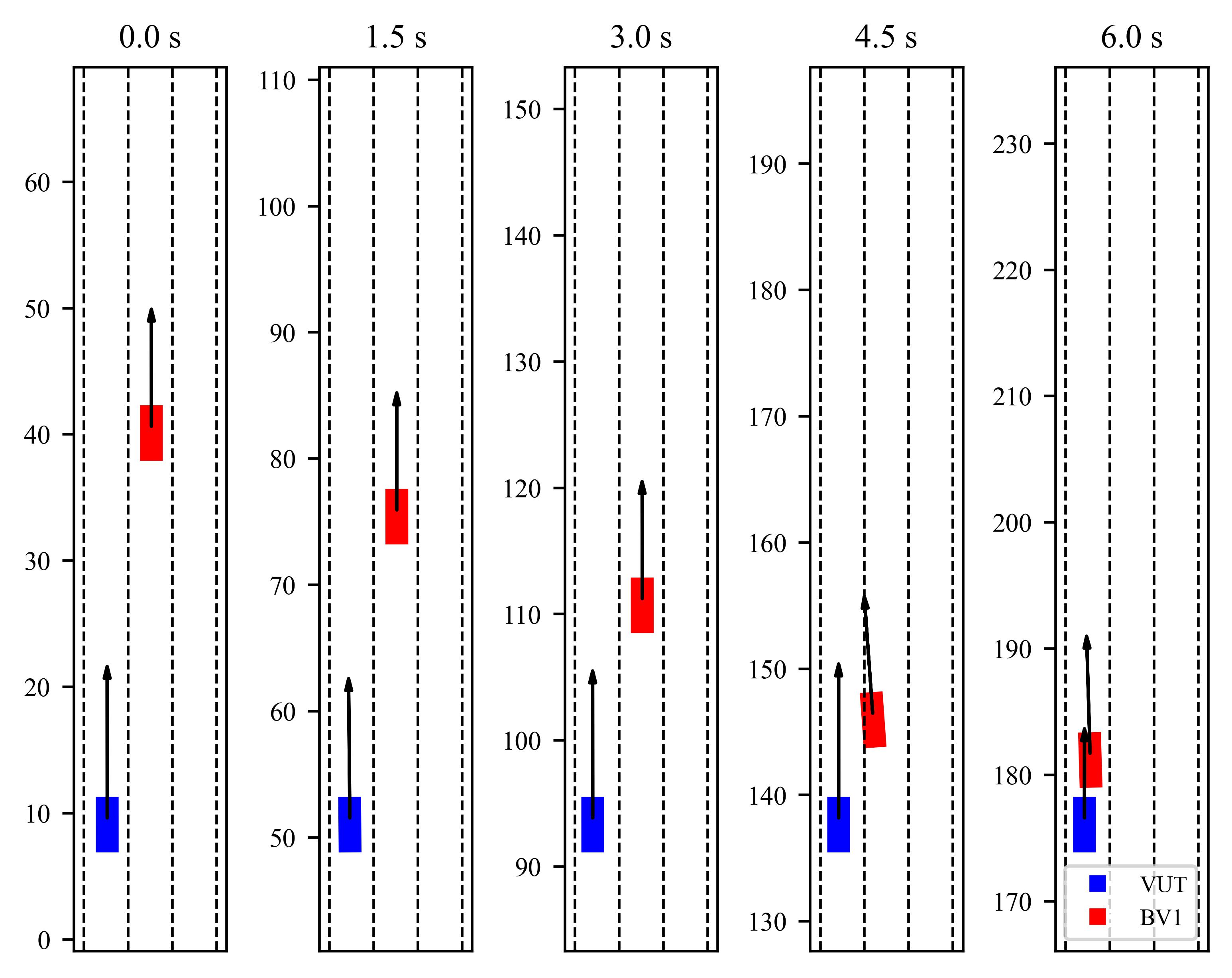}
        }
    \subfigure[Hazardous Scenario 2. $(S_1=48.72m,V_1=20.03m/s,T_{lc}=2.02s)$]{
        \label{sce2636}
        \includegraphics[width=\linewidth]{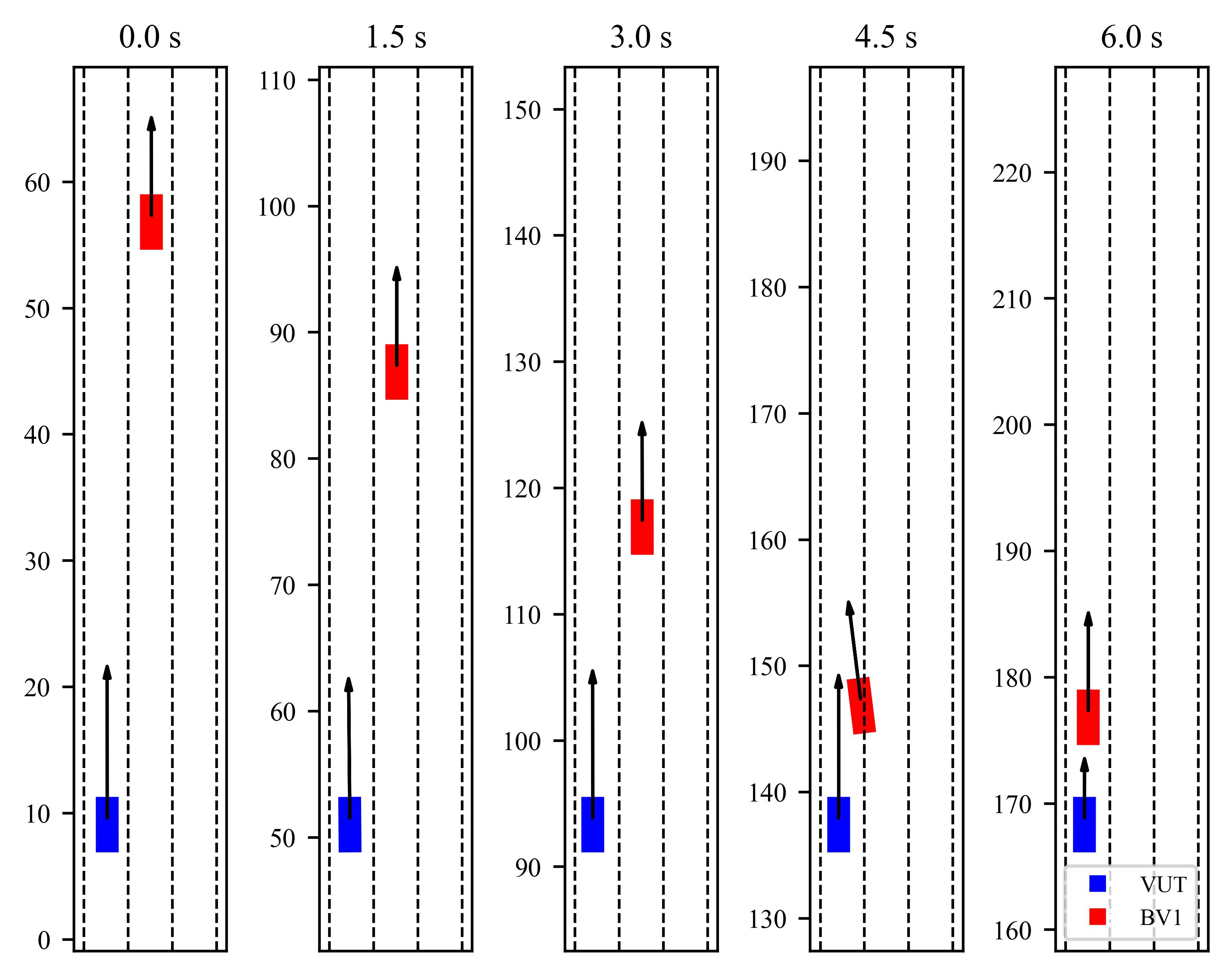}
        }
    \caption{Visualization of the two selected hazardous scenarios.
    }
    \label{sce_visual}
\end{figure}

As shown in Fig. \ref{3dScenario_res}, it is evident that the points searched by the algorithm are distinctly clustered within and in the vicinity of the hazardous area as shown in Fig. \ref{3dScenario_GT}, demonstrating the algorithm's capacity to accelerate the testing process. Secondly, all the searched hazardous scenarios are identified as one hazardous domain. In order to analyze the similarities and differences of different hazardous scenarios within the same hazardous domain, we select the two most distant hazardous scenarios within the hazardous domain (namely the Hazardous scenario 1 and 2 labeled in Fig. \ref{3dScenario_res}) for a case study. The two scenarios are visualized in Fig. \ref{sce_visual}, with the length of the arrow representing the velocity of the vehicles.

By observing Fig. \ref{sce_visual}, we can find that although the initial conditions are different in the two scenarios, the hazards occur for the same reason. Specifically, in both scenarios, BV1 starts to change lanes at 3s, and then at 4.5s, BV1 starts to/has entered the VUT's lane. However, at this moment, VUT does not execute any evasive maneuvers and remains in its original speed. Actually, it is not until BV1 almost completes its lane changing maneuver that VUT begins to brake. The delayed response of the VUT is inferred to be the cause of the hazards in both scenarios. The above analysis leads us to the conclusion: Different hazardous scenarios within the same hazardous domain share the same hazard generation mechanism. Therefore, when evaluating the safety performance of VUTs, it is sufficient to analyze only one scenario in each hazardous domain, rather than analyzing each hazardous scenario individually, which significantly enhances the efficiency of safety evaluation.

\subsection{Stopping Condition Validation} \label{sec_stop_condi}

\begin{figure}[t]%
    \centering
    \subfigure[The stopping condition coverage indicator $C_T$]{
        \label{stop_ct}
        \includegraphics[width=8cm]{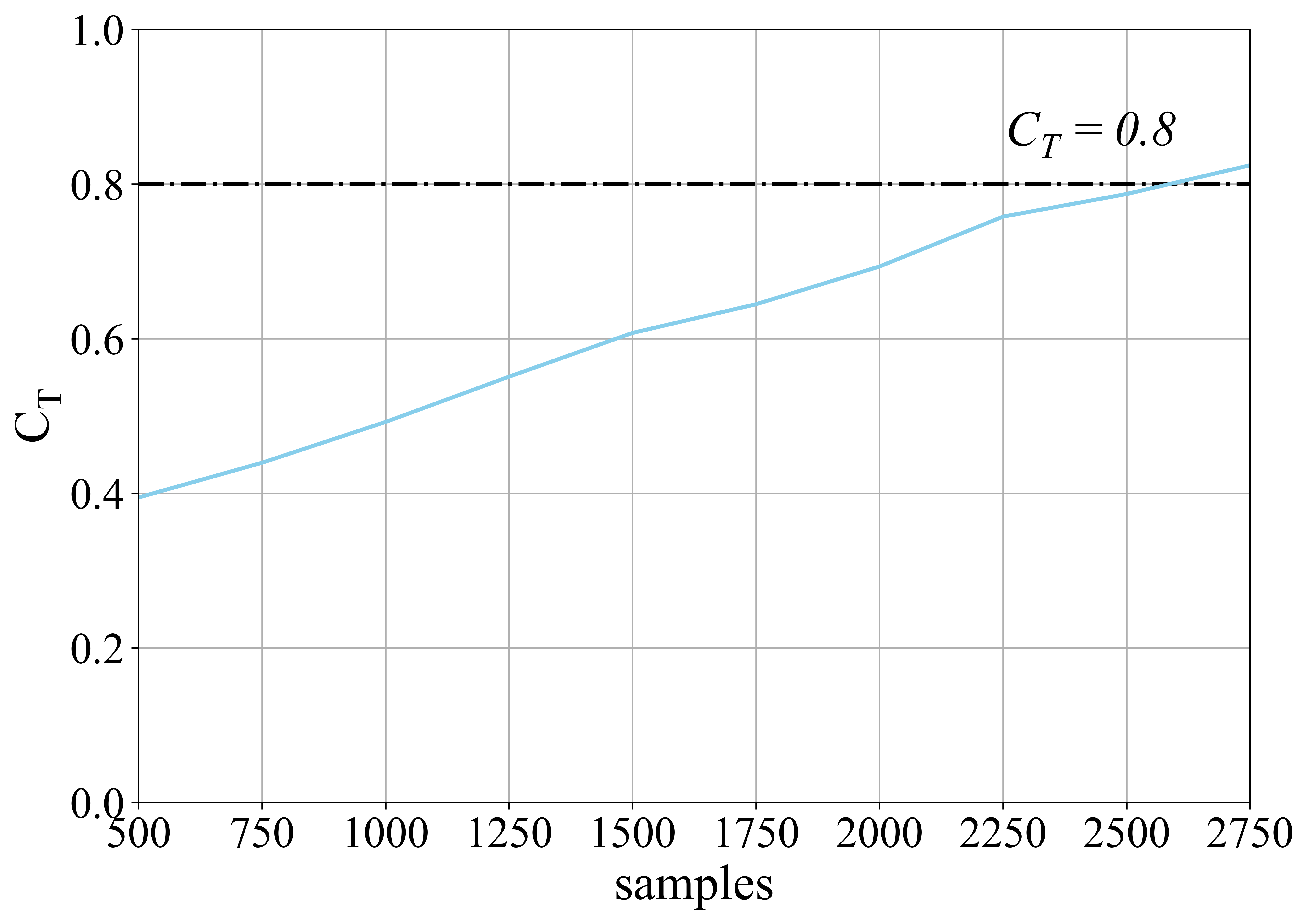}
        }
    \subfigure[The stopping condition predictive performance indicator $F_{2-obv}$, with $F_{2-grid}$ as a reference]{
        \label{stop_f2}
        \includegraphics[width=8cm]{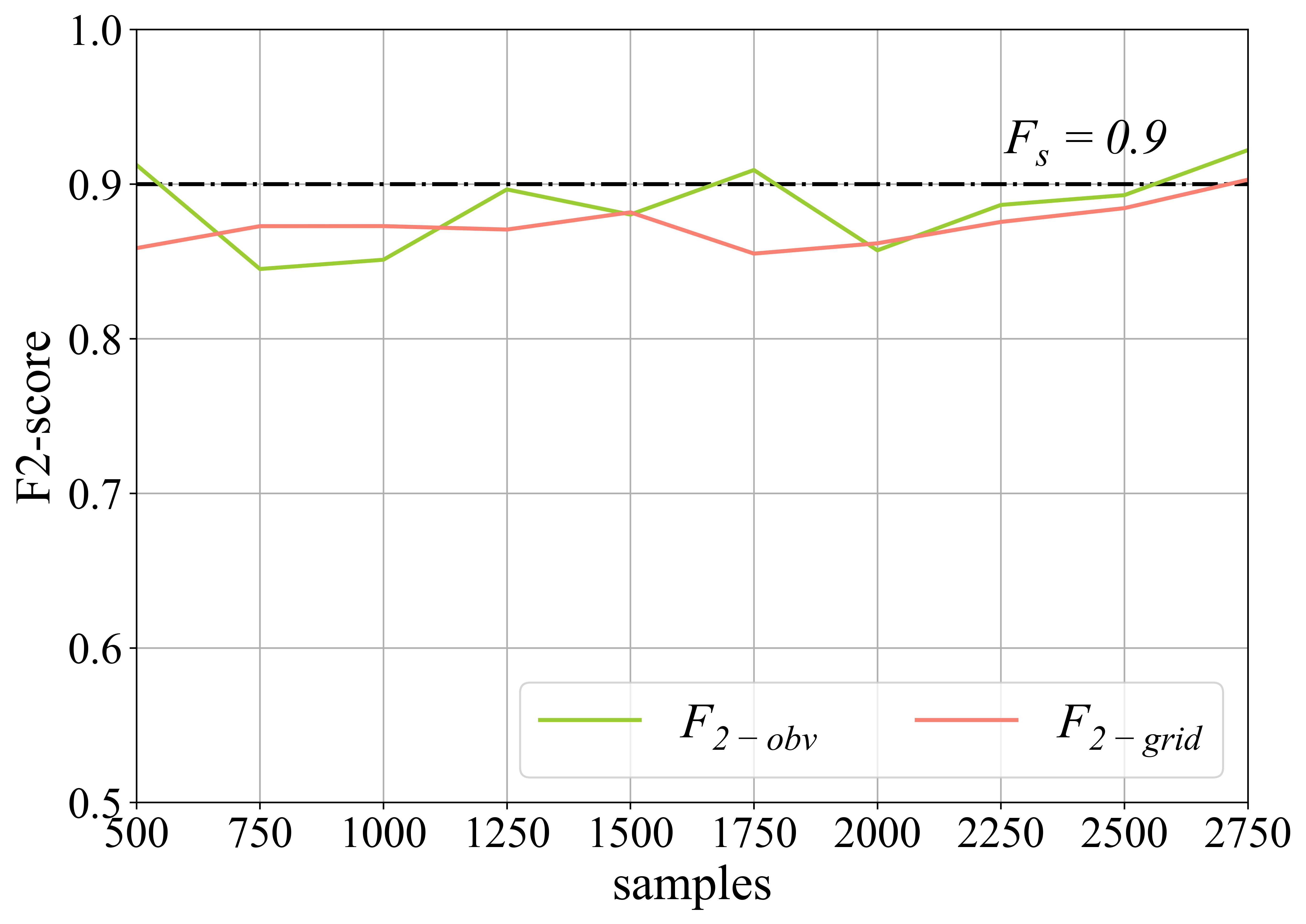}
        }
    \caption{Stopping Condition Validation.
    }
    \label{stop_condi_valid}
\end{figure}

The stopping condition proposed in this paper is used in the 3-d cut-in scenario. Specifically, we perform the stopping condition checking after 500 samples because the initial sampling is sparse. Then the stopping condition is checked every 250 samples. During the above process, the two stopping indicators $C_T$ and $F_{2-obv}$ are examined in each round to determine whether the search should stop or not. Their trend with the number of samples is shown in Fig. \ref{stop_condi_valid}. Furthermore, the $F_{2-grid}$ is also given as a reference to verify that the search algorithm stops at the correct time.

As shown in Fig. \ref{stop_condi_valid}, the threshold $F_s$ is chosen as 0.9 in this experiment. When the search proceeds to 2750 samples, both $C_T$ and $F_{2-obv}$ satisfy the stopping requirements and the search stops. At this moment, $F_{2-grid}$ exactly reaches 0.9. Therefore, the stopping condition proposed in this paper ensures the adequacy of the search without wasting additional computational resources. Additionally, it can be seen in Fig. \ref{stop_condi_valid} that $F_{2-obv}$ exceeds 0.9 several times before the search algorithm finally stops. However, at those points, $F_{2-grid}$ still maintains a low value, indicating an inadequate search and the stopping condition is not triggered due to the low $C_T$, which shows again the reasonableness of the stopping condition setting in this paper.

\section{Conclusion} \label{conclusion}

This paper proposes an integrated accelerated testing and evaluation method (ITEM) for ADSs with the goal to make full use of the testing information. Based on a Monte Carlo tree search paradigm and a dual surrogates testing framework proposed in our previous work, this paper considers the testing stage and the evaluation stage as a whole and bridges the gap between the two stages. Specifically, to serve the purpose of accurate hazardous domains identification in the evaluation stage, we first modify the UCB calculation method to focus more on the boundary subspaces that contain both hazardous and safe scenarios. By doing this the hazardous boundaries are carefully explored. Next, we propose a hazardous domain identification method using both the sampled points and the tree structure (which contains information of the affiliation of each historical sampled point with the subspaces and the parent-child relationship between subspaces) to obtain rectangular hazardous domains. After that, to validate the proposed method, we propose two evaluation metrics (namely, API and ADI) to calculate the accuracy of the proportional and distributional identification. Experimental results show that our method outperforms other baseline methods in terms of accuracy and generality. And the efficiency of the improvements is also verified through ablation experiments. Additionally, we construct a stopping condition based on the convergence of the optimization algorithm to enable self-stopping of the testing so that the computational resources will not be wasted.

Since both hazardous scenario ratio estimation and hazardous domain identification are included our framework, our future work will focus on the hazardous scenario ratio computation within each identified hazardous domain, which will lead to a more precise and comprehensive evaluation of the safety performance of ADSs. At the same time, more advanced ADSs in higher dimensional scenarios will be carried out to demonstrate the practical value of our proposed method.

\appendix
\section{Appendix} \label{appendix}

The calculations of $V_{exploit}$, $\bar{\epsilon}_{Node}$ and $\bar{\rho}_{Node}$ are essentially the same as in \cite{wu2024accelerated} and \cite{wu2024lambda} with minor modifications. For ease of understanding, here we give a summary of the calculation formulas for all these parameters.

The exploitation value of the child node $B$ $V_{exploit}$  is calculated as:

\begin{equation}\label{Apend_eq1}
\begin{aligned}
V_{exploit} = \sum_{\boldsymbol{x}\in \mathcal{D}_B} f(\boldsymbol{x})w_B(\boldsymbol{x})
\end{aligned}
\end{equation}
In which,

\begin{equation}\label{Apend_eq2}
\begin{aligned}
w_B(\boldsymbol{x}) = \frac{1/\rho(\boldsymbol{x})}{\sum_{\boldsymbol{x}\in \mathcal{D}_B} 1/\rho(\boldsymbol{x})}
\end{aligned}
\end{equation}
where $\mathcal{D}_B$ is the set of all sampled points in node $B$. $f(\boldsymbol{x})$ is the risk result calculated by the safety metric. $w_B(\boldsymbol{x})$ is the density weight of the sampled point $\boldsymbol{x}$. $\rho(\boldsymbol{x})$ is the density at $\boldsymbol{x}$, which is estimated by Kernel Density Estimator (KDE).

With the density weight of each sampled point known, the average density $\bar{\rho}_{Node}$ and average loss $\bar{\epsilon}_{Node}$ of a certain node can be obtained by Eq. \ref{Apend_eq3}.

\begin{equation}\label{Apend_eq3}
\begin{aligned}
\bar{\rho}_{Node} =\sum_{\boldsymbol{x}\in \mathcal{D}_B} w_{Node}(\boldsymbol{x})\rho(\boldsymbol{x}) \\
\bar{\epsilon}_{Node} = \sum_{\boldsymbol{x}\in \mathcal{D}_B} w_{Node}(\boldsymbol{x})e(\boldsymbol{x})
\end{aligned}
\end{equation}
where $e(\boldsymbol{x})$ is the loss function of the behavior surrogate model.

\printcredits

\bibliographystyle{cas-model2-names}

\bibliography{citelist}

\end{sloppypar}
\end{document}